\documentclass{article}

\usepackage{microtype}
\usepackage{graphicx}
\usepackage{subfigure}
\usepackage{booktabs} 
\usepackage{subcaption}

\usepackage{hyperref}

\usepackage[accepted]{icml2024}

\usepackage{amsmath}
\usepackage{amssymb}
\usepackage{mathtools}
\usepackage{amsthm}

\usepackage[capitalize,noabbrev]{cleveref}

\theoremstyle{plain}

\theoremstyle{definition}

\theoremstyle{remark}

\usepackage[textsize=tiny]{todonotes}

\usepackage{soul}

\icmltitlerunning{Attamba: Attending To Multi-Token States}

\begin{document}
\twocolumn[
\icmltitle{Attamba: Attending To Multi-Token States}




\begin{icmlauthorlist}
\icmlauthor{Yash Akhauri}{cornell,google}
\icmlauthor{Safeen Huda}{google}
\icmlauthor{Mohamed S. Abdelfattah}{cornell}
\end{icmlauthorlist}

\icmlaffiliation{cornell}{Cornell University}
\icmlaffiliation{google}{Google}

\icmlcorrespondingauthor{Yash Akhauri}{ya255@cornell.edu}

\icmlkeywords{Machine Learning, ICML}

\vskip 0.3in
]



\printAffiliationsAndNotice{} 


\begin{abstract}

When predicting the next token in a sequence, vanilla transformers compute attention over all previous tokens, resulting in quadratic scaling of compute with sequence length. 
State-space models compress the entire sequence of tokens into a fixed-dimensional representation to improve efficiency, while other architectures achieve sub-quadratic complexity via low-rank projections or sparse attention patterns over the sequence.
In this paper, we introduce Attamba, a novel architecture that uses state-space models to compress chunks of tokens and applies attention on these compressed key-value representations. 
We find that replacing key and value projections in a transformer with SSMs can improve model quality and enable flexible token chunking, resulting in 24\% improved perplexity with transformer of similar KV-Cache and attention footprint, and $\approx 4\times$ smaller KV-Cache and Attention FLOPs for 5\% perplexity trade-off. 
Attamba can perform attention on chunked-sequences of variable length, enabling a smooth transition between quadratic and linear scaling, offering adaptable efficiency gains. \href{https://wandb.ai/akhauriyash/attamba_arxiv}{[Logs]} \href{https://github.com/abdelfattah-lab/attamba}{[Code]}
\end{abstract}


\begin{figure}[ht!]
    \centering
    \includegraphics[width=1.03\linewidth]{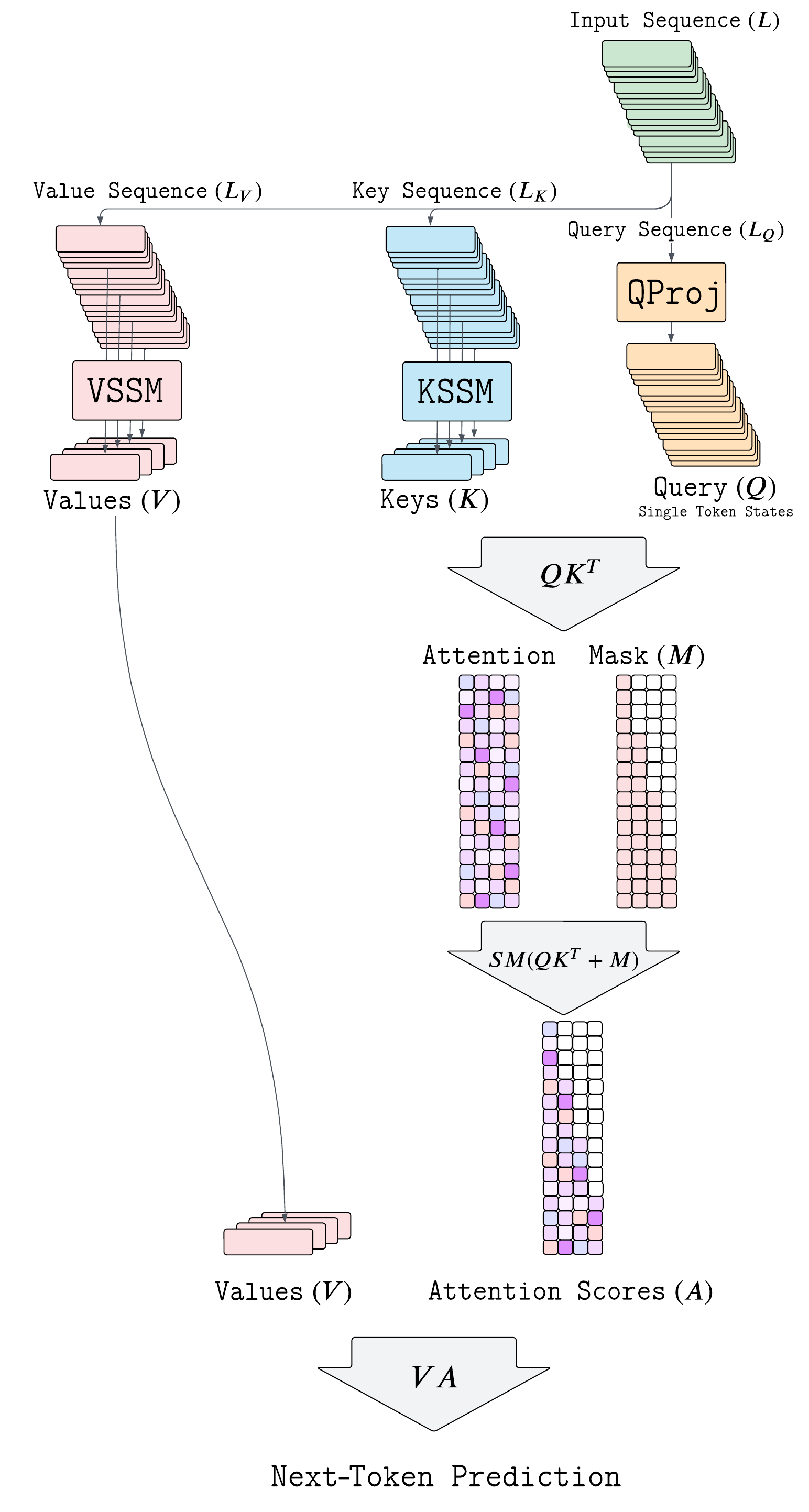}
    \caption{Attamba uses State-Space Models (SSM) to compress key-value sequences into token chunks (e.g., chunks of $P = 4$), reducing the attention map and KV-Cache size  by $P\times$ by storing only chunk boundaries.}
    \label{fig:attamba_teaser}
\end{figure}

\section{Introduction}
\label{sec:intro}



Transformers have provided an effective and scalable sequence modeling architecture, leading to major strides in natural language processing. 
This comes at a high cost when processing long sequences due to the quadratic complexity of attention. 
Efforts to ameliorate this inefficiency have faced challenges for tasks requiring extended contexts, as dropping tokens that may later need to be referenced can render many token-pruning techniques \cite{zhang2023h2o,xiao2024efficientstreaminglanguagemodels} ineffective. 
To address the inefficiency of standard attention, several approaches have been developed. KV-Cache compression techniques, such as Palu\cite{chang2024palucompressingkvcachelowrank} uses low-rank projections to compress hidden dimensions, while ShadowKV~\cite{sun2024shadowkvkvcacheshadows} uses low-rank key caching for long-context inference. 
Methods such as LinFormer \cite{wang2020linformer} and PerFormer \cite{choromanski2020performer} use low-rank approximations or kernel-based projections to reduce complexity of attention.
Sparse attention models such as BigBird \cite{zaheer2020bigbird} adopt fixed attention patterns, but these can fail in settings where static sparsity may not capture necessary interactions. 

In contrast, State-Space Models (SSMs) \cite{gu2021efficiently_ssm1, gu2020hippo_ssm2} including architectures like Mamba \cite{gu2023mamba, dao2024mamba2} compress entire sequence histories into fixed-dimensional states, offering linear complexity. 
However, SSMs face challenges in representing arbitrarily long contexts with the same expressivity as the attention mechanism. 
Stuffed Mamba \cite{chen2024stuffedmamba} highlights the phenomenon of \textit{state collapse}, which arises when the recurrent state of RNN-based architectures like Mamba fail to generalize to sequences longer than those seen during training \cite{wang2023stablessm_statecollapse, merrill2024illusion_statecollapse}. 
Despite efficient memory use, the fixed-dimensional state of SSMs has an upper bound on information representation, which once exceeded, cannot effectively retain earlier contextual information. 

\begin{figure}[ht!]
    \centering
    \includegraphics[width=\linewidth]{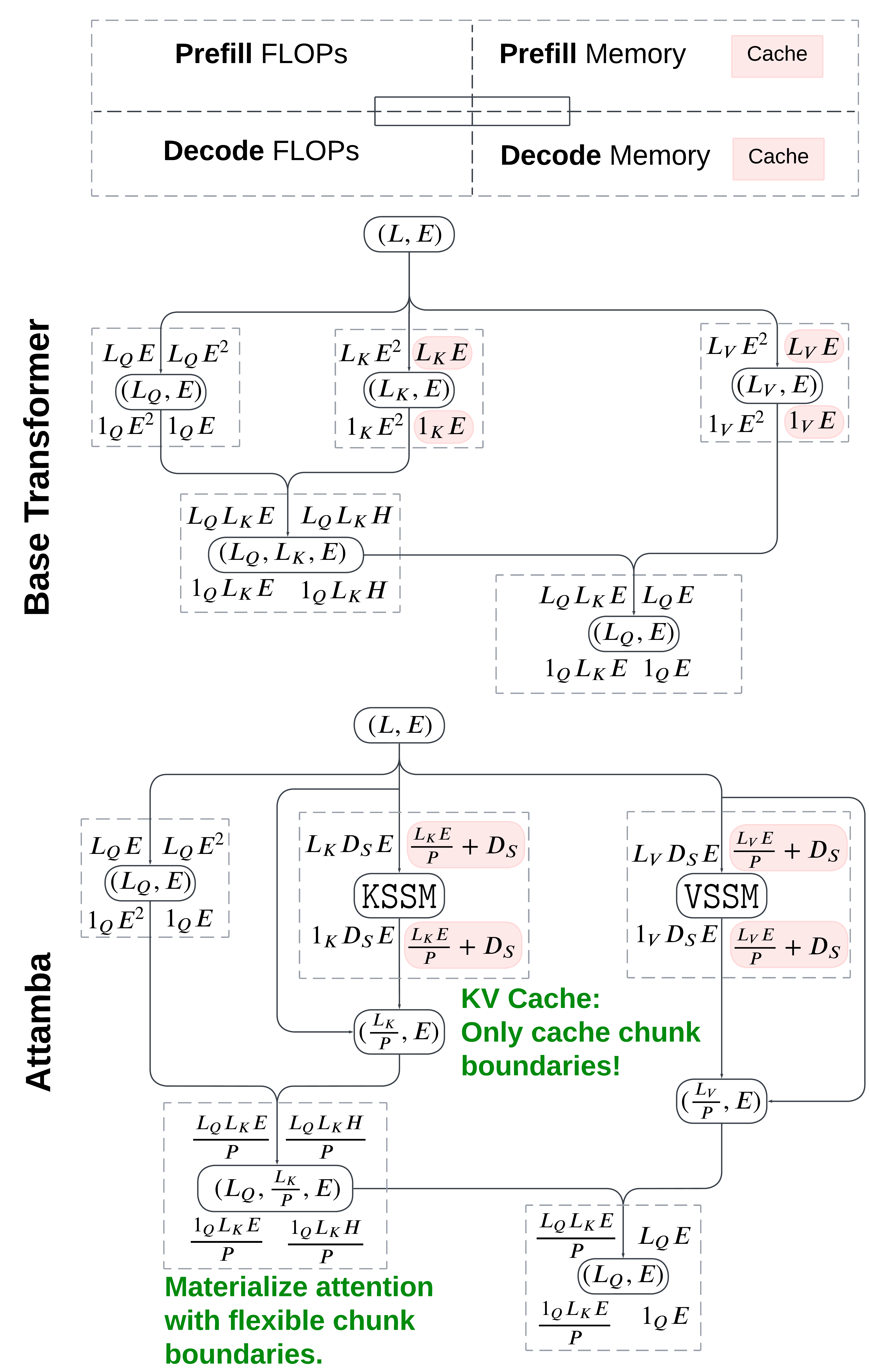}  
    \caption{State Space Models (SSMs) efficiently encode multiple tokens into a single representation. By compressing key (\( K \)) and value (\( V \)) sequences into chunked representations, SSMs maintain essential contextual information, enabling efficient query (\( Q \)) interactions. This approach minimizes KV-Cache size by storing only chunk boundaries and reduces the computational cost of attention. Attamba demonstrates robustness to randomized chunk boundaries, indicating the potential for flexible computation-quality trade-offs. Approximate FLOPs/Memory shown, constants  ignored. Variables: \( L \) (Sequence length), \( P \) (Chunk size), \( D_{S} \) (SSM state dimension), \( E \) (Model dimension).}
    \label{fig:attamba_complexity}
\end{figure} 

Our key insight is that this limitation can be leveraged: SSMs can be adapted to learn how to compress chunks of tokens into meaningful, single-token states that preserve essential information. 
By training SSMs to perform variable-length token chunking, SSMs can consolidate sequences of tokens into compact representations, which can then be processed by the standard attention mechanism, reducing the $L^{2}$ attention operations by a factor of chunk size. In this paper, we introduce \textit{Attamba}, a novel architecture that combines SSMs and Attention. 
As shown in Figure \ref{fig:attamba_teaser}, Attamba uses SSMs for key-value projections, enabling flexible \textit{chunked attention} by compressing multiple tokens into a single state. 
As illustrated in Figure \ref{fig:attamba_complexity}, Attamba significantly reduces KV-cache memory and attention computation costs by caching only the compressed chunk boundaries, rather than all tokens. Our contributions are:

\begin{itemize}
    \item \textbf{Integration of SSMs into Attention:} By replacing key-value projection matrices with SSM blocks, we demonstrate that it is possible to compress multiple tokens into one representation, while effectively attending to these representations.
    \item \textbf{Efficient Token Chunking}: We introduce cyclic token chunking when compressing multiple tokens to a single SSM state to reduce bias from fixed boundaries. We also demonstrate the  feasibility of variable-length token chunking and present over $8\times$ KV-cache and attention savings with minimal perplexity trade-offs.
\end{itemize}


\section{Related Work}
\label{sec:rel_work}

\textbf{Attention: }
Transformers are foundational for language modeling but face challenges due to the quadratic complexity of attention, which grows with the square of the sequence length. This makes attention computation both memory-intensive and computationally expensive. Additionally, during auto-regressive inference, the key-value cache size grows linearly with the sequence length and embedding dimension, adding significant memory overhead. These factors limit efficiency and scalability, especially for long-context applications. Efforts to mitigate this inefficiency include \textit{LinFormer} \cite{wang2020linformer}, which reduces complexity via low-rank factorization (\texttt{k}), and \textit{BigBird} \cite{zaheer2020bigbird}, which uses sparse attention patterns (\texttt{r,w,g} denoting random, window, global tokens) to handle long sequences more efficiently. \textit{PerFormer} \cite{choromanski2020performer} leverages kernel-based approximations to achieve sub-quadratic complexity. While effective, these methods face limitations in preserving attention expressivity, especially in long-context tasks.. 

\textbf{State-Space Models: }
State-Space Models provide an efficient mechanism for long-sequence processing. \textit{Mamba} \cite{gu2023mamba} and its successor \textit{Mamba2} \cite{dao2024mamba2} are notable implementations that achieve linear complexity by compressing sequence history into fixed-dimensional states. However, these models struggle with information retention over arbitrarily long contexts, as discussed in \textit{Stuffed Mamba} \cite{chen2024stuffedmamba} which highlights the state collapse issue. 

\textbf{Hybrid Models: }
Combining the strengths of attention and SSMs, hybrid models have emerged. \textit{Jamba} \cite{lieber2024jamba} interleaves Transformer and Mamba layers, using a mixture-of-experts (MoE) approach to manage parameter usage and support long-context modeling efficiently. \textit{Griffin} \cite{de2024griffin} integrates gated linear recurrences with local attention, achieving efficient scaling and superior performance on extrapolation tasks. Similarly, \textit{Hawk} \cite{de2024griffin} utilizes recurrent blocks to outperform Mamba on various downstream tasks. Techniques like \textit{Multi-Token Prediction (MTP)} \cite{gloeckle2024multitokenpred} optimize efficiency by predicting multiple tokens simultaneously, improving sample efficiency and enabling faster inference.
Hybrid approaches like \textit{Samba} \cite{ren2024samba} and \textit{Jamba} explore novel trade-offs between efficiency and expressivity. \textit{Samba} employs sliding-window attention combined with state-space layers. Our approach differentiates itself by directly integrating SSM blocks inside the attention mechanism instead of interleaving SSMs and Transformer blocks.

\section{Preliminaries}
\subsection{Attention}


Let \(\mathbf{X} \in \mathbb{R}^{n \times e}\) be the input to the attention mechanism, where \(n\) is the sequence length and \(e\) is the model embedding dimension. The embedding dimension \(e\) can be expressed as \(e = h \times d\), where \(h\) is the number of attention heads, and \(d\) is the per-head dimension. The projection matrices \(W^Q, W^K, W^V \in \mathbb{R}^{e \times e}\) are used to compute the query, key, and value representations respectively. \(\texttt{SM}\) denotes the softmax operation, and \(\texttt{Attn}\) represents the attention computation. \(S\) represents the scaled attention scores, and \(A\) represents the attention probabilities (normalized weights).

\begin{equation}
\texttt{Attn}(\mathbf{X}) = \underbrace{\texttt{SM}\left( \underbrace{\frac{\mathbf{X} W^Q (\mathbf{X} W^K)^T}{\sqrt{d}}}_{S} \right)}_{A} \cdot \mathbf{X} W^V
\label{eqn:attnmechanism}
\end{equation}

\subsection{SSMs}
State Space Models maintain a hidden state $\mathbf{x(t)} \in \mathbb{R}^{D_{S}}$, which evolves over time based on the input sequence and state transition matrices. Computing the output sequence from a given input is linear in complexity, requiring $\mathcal{O}(nD_{S})$ in time-complexity and $\mathcal{O}(D_{S})$ in space. In the Mamba framework \cite{gu2023mamba}, variable-length sequence handling is streamlined using \texttt{cu\_seqlens} (\textbf{c}umulative \textbf{u}nique \textbf{se}quence \textbf{len}gths), which denotes cumulative sequence lengths. This allows efficient indexing of flattened batch sequences, avoiding padding overhead. We leverage \texttt{cu\_seqlens} for efficient processing of chunked-sequences. 
In Section \ref{subsec:var_tok_chunk}, we investigate different schemes for token chunking.

\begin{figure}[bp!]
    \centering
    \includegraphics[width=0.8\linewidth]{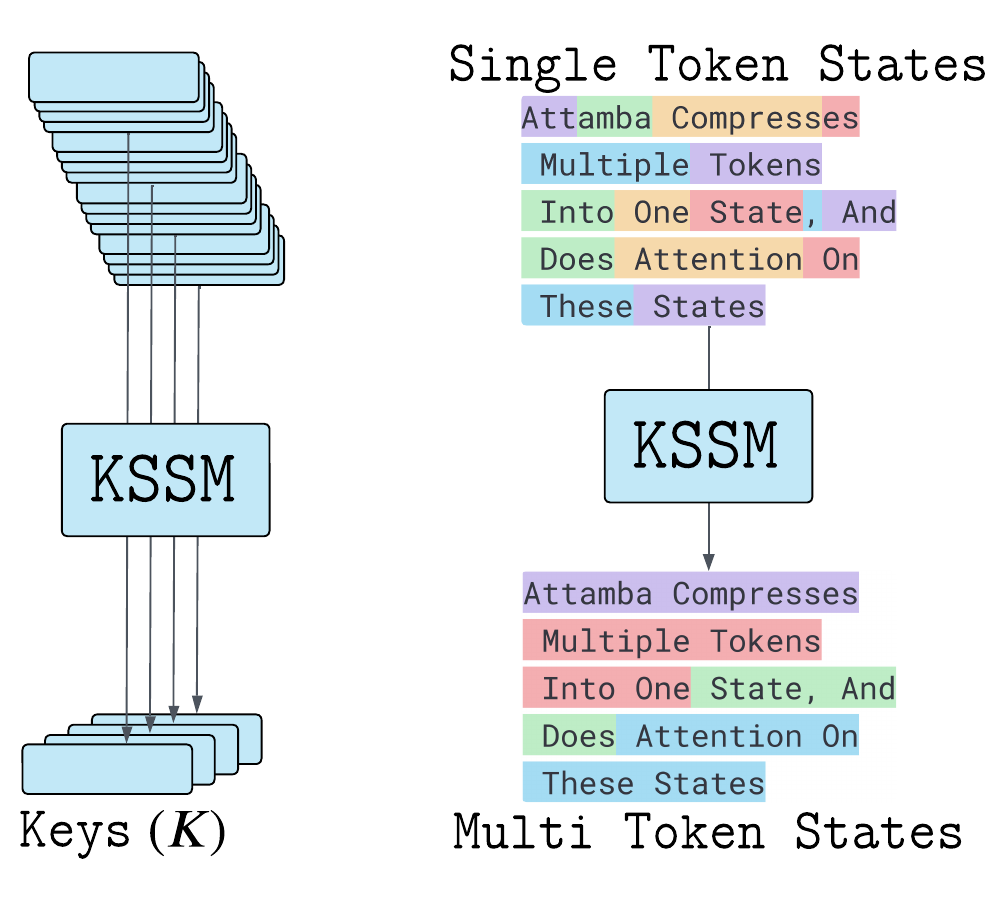}
    \caption{Attamba uses SSM blocks to compress chunks of tokens ($P = 4$ in the example above) into a single token.}
    \label{fig:chunkdemo}
\end{figure}

\begin{figure*}[ht!]
    \centering
    \includegraphics[width=\linewidth]{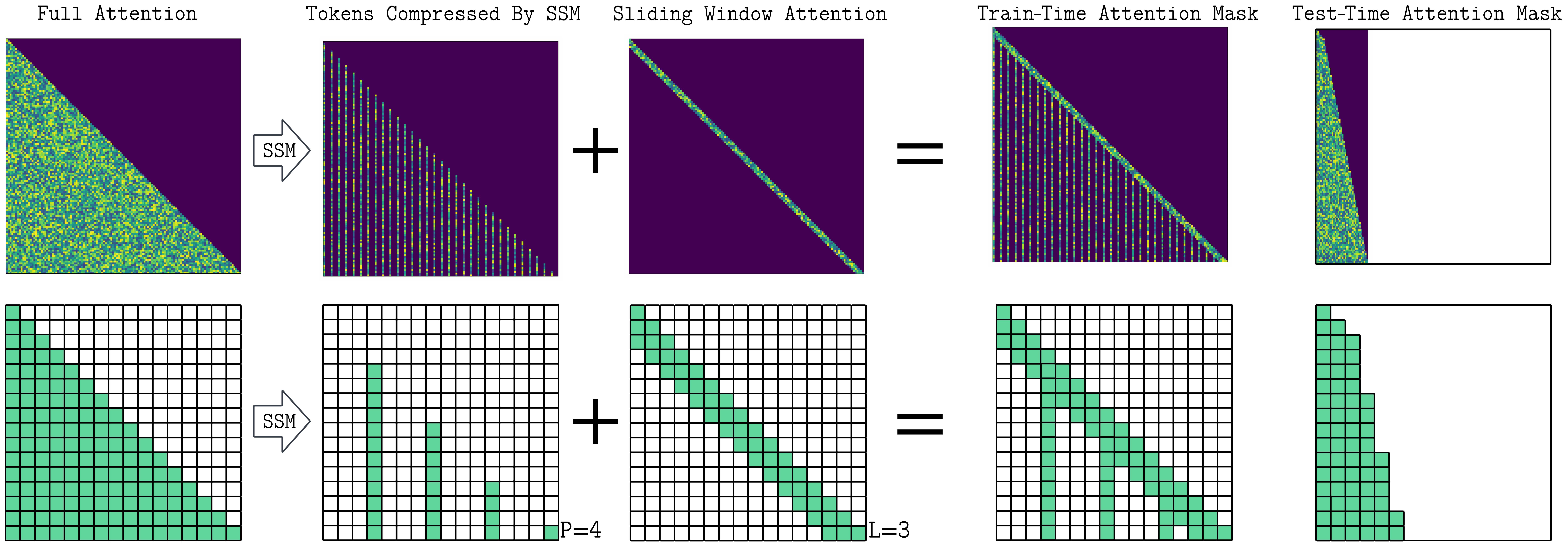}
    \caption{Full-Attention has a purely causal mask, attending to all past tokens. Attamba uses Key-Value SSM blocks to compress chunks of $P$ tokens (e.g. $P=4$) into one state. Tokens compressed by SSMs are at chunk boundaries. This is incorporated with a sliding-window attention (when $L > 1$). At test-time (inference), only the chunk boundaries and sliding window tokens need to be preserved, reducing KV-Cache and Attention FLOPs. }
    \label{fig:attamba_algo}
\end{figure*}

\subsection{Auto-regression and Masking}
\label{subsec:ar_and_mask}

Transformers are trained for next-word prediction, by modelling the probability of each token, given all previous tokens in a sequence. To enforce \textit{causality} (to not attend to future tokens), a causal mask is applied to the attention mechanism during training, when computing $\mathbf{A}$ in Equation \ref{eqn:attnmechanism}. Specifically, the causal mask $M \in \mathbb{R}^{n \times n}$ is defined to prevent positions from attending to future tokens as below:

\begin{equation}
M_{i,j} = 
\begin{cases} 
0, & \text{if } j \leq i, \\ 
-\infty, & \text{if } j > i. 
\end{cases} 
\label{eqn:causal_mask}
\end{equation}

This $M$ mask is applied before the softmax, resulting in $\mathbf{A} = \texttt{SM}(S + M)$. Setting elements of $M$ to $-\infty$, the attention weights become zero after the softmax, which effectively excludes future tokens from the computation. This mechanism can also be used to emulate a token being omitted by appropriately adjusting the mask. 

In next-word prediction tasks, the output at position $k$ depends on all previous tokens from positions $0$ to $k-1$, building up cumulative information in hidden states, with each position $k$ capturing knowledge of all tokens up to that point. This property is essential for capturing dependencies across the sequence. Leveraging this cumulative information property of auto-regressive models such as SSMs and Transformers, along with the flexibility of mask $M$ controlling token omissions makes it possible to control range and choice of tokens each position attends to.

\section{Attamba: Attentive SSMs}

Auto-regressive transformers and SSMs enable us to compress information about prior tokens into a singular, final representation. 
For next-word prediction, this property is used to transform the representation to what the next word should be. 
Further, we can control what past information transformers attend to with the attention mask, described in Section \ref{subsec:ar_and_mask}. Attamba leverages these properties to \textbf{(1)} compress $P$ tokens into a single token using SSM blocks, exhibiting linear complexity, and \textbf{(2)} leverage attention mask to attend to only these compressed states for efficient training and inference. Specifically, we preserve the query sequence length, to enable \textit{causally valid} training of all next-word prediction problems in a given input sequence, and replace the key and value projection matrices with SSM blocks.

\subsection{Formulation}
Attamba integrates State Space Models (SSMs) into the attention mechanism to efficiently handle long sequences. As seen in Figure \ref{fig:attamba_complexity}, it replaces key and value projection matrices with SSM blocks and a residual connection that processes chunks of tokens, reducing computational complexity of attention while preserving context of input sequence. 

Let $P$ denote the chunk size, i.e., the number of tokens processed by the SSM at a time. 
Given an input sequence $\mathbf{X} \in \mathbb{R}^{n \times e}$, the query vector is computed as usual to preserve the auto-regressive nature of transformers. 
However, the keys and values are obtained by processing the input sequence through SSMs. 
The sequence is divided into non-overlapping chunks of size $P$ (Figure \ref{fig:chunkdemo}), and each chunk is processed auto-regressively by the SSM. 
For simplicity, we assume $n$ is divisible by the chunk size $P$ in this discussion, though SSMs can seamlessly handle partial chunks, as they do during auto-regressive inference in Attamba.

Let $\mathbf{X}^{(p)} \in \mathbb{R}^{P \times e}$ denote the $p$-th chunk of the input sequence, where $p = 1, 2, \dots, \frac{n}{P}$. The SSM processes each chunk to produce compressed key and value representations:

\begin{equation} 
\begin{aligned} \mathbf{K}^{(p)} &= \text{SSM}_K\left(\mathbf{X}^{(p)}\right), \ \mathbf{V}^{(p)} &= \text{SSM}_V\left(\mathbf{X}^{(p)}\right)\end{aligned} 
\label{eqn:ssm_to_kv}
\end{equation}

where $\text{SSM}_K$ and $\text{SSM}_V$ denote the SSMs used for keys and values, respectively.

At \textbf{train-time}, we need to preserve all SSM outputs, since next-word prediction problems require attending to incomplete (partial) chunks as well. Thus, we keep the SSM outputs for every token, giving us:

\begin{equation} 
\begin{bmatrix} \mathbf{K}_{\text{SSM}} \\ \mathbf{V}_{\text{SSM}} \end{bmatrix} = 
\begin{bmatrix} 
\mathbf{K}^{(1)} & \mathbf{K}^{(2)} & \cdots & \mathbf{K}^{(n)} \\ 
\mathbf{V}^{(1)} & \mathbf{V}^{(2)} & \cdots & \mathbf{V}^{(n)} 
\end{bmatrix} \in \mathbb{R}^{2 \times n \times e}
\end{equation}

To perform attention, the queries $\mathbf{Q}$ attend to compressed keys-values at chunk boundaries and the latest partial chunk (Self-Attention). Thus, the attention mask $M_{\text{train}}$ must account for both causality and chunk boundaries:

\begin{equation} 
\left(M_{\text{train}}\right)_{i,j} = 
\begin{cases} 
0, & \text{if } \left( \left\lfloor \dfrac{j}{P} \right\rfloor = \left\lfloor \dfrac{i}{P} \right\rfloor \text{ and } j \leq i \right) \\
& \quad \text{or} \ \left( j \leq i \text{ and } j \bmod P = P - 1 \right), \\
-\infty, & \text{otherwise}.
\end{cases}
\end{equation}

At \textbf{test-time}, the outputs $\mathbf{K}^{(p)}[-1], \mathbf{V}^{(p)}[-1] \in \mathbb{R}^{1 \times e}$ are compressed representations of each chunk, as only the final representation is needed. This is obtained by taking $(\mathbf{K}^{(p)}[-1], \mathbf{V}^{(p)}[-1])$ from Equation \ref{eqn:ssm_to_kv}. By concatenating these, we obtain:

\begin{equation} 
\vspace{-3mm}
\begin{bmatrix} \mathbf{K}_{\text{SSM}} \\ \mathbf{V}_{\text{SSM}} \end{bmatrix} = 
\begin{bmatrix} 
\mathbf{K}^{(1)}[-1] &  \cdots & \mathbf{K}^{(\frac{n}{P})}[-1] \\ 
\mathbf{V}^{(1)}[-1] &  \cdots & \mathbf{V}^{(\frac{n}{P})}[-1] 
\end{bmatrix} \in \mathbb{R}^{2 (\frac{n}{P}) \times e}
\end{equation}

As shown in Figure \ref{fig:attamba_algo}, at test-time, an appropriate attention mask $M_{\text{test}}$ can be constructed:

\begin{equation} 
\left(M_{\text{test}}\right)_{i,j} = 
\begin{cases} 
0, & \text{if } j \leq \left\lfloor \frac{i}{P} \right\rfloor , \\ 
-\infty, & \text{otherwise}.
\end{cases}
\end{equation}

By replacing key and value projections with SSMs that compress chunks of tokens, we reduce the computational cost of the attention mechanism from $(n^{2}e)$ to $(n^{2}e/P)$. We can also achieve $\mathcal{O}(nPe)$ complexity if we divide the input sequence length into P chunks, irrespective of sequence length, resembling Attamba-Linear in Figure \ref{fig:attamba_variants}. 
We find that SSMs are robust to even randomized chunk boundaries, which may facilitate this complexity trade-off.

\begin{figure}[bh!]
\vspace{5mm}
    \centering
    \includegraphics[width=\linewidth]{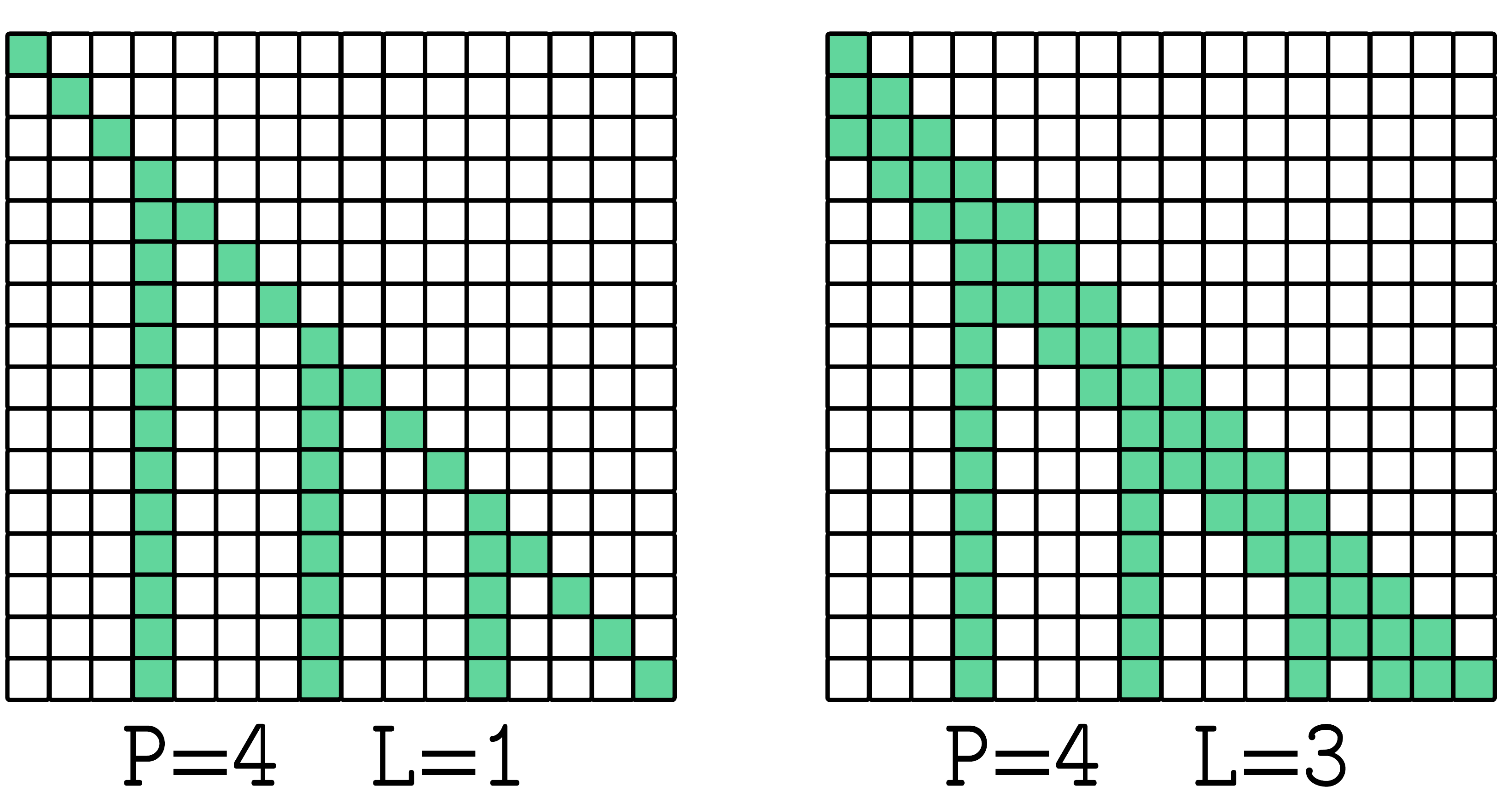}
    \caption{Leading-Tokens (\textbf{L}) control how many 'leading' tokens full-attention happens over, preserving full-attention on the newest tokens. This resembles Sliding-Window attention. Chunk-size (\textbf{P}) controls how many tokens are chunked by the SSM.}
    \label{fig:leading_chunking}
\end{figure}

\textbf{Chunk Sizes and Leading Tokens: } In Attamba, the notion of \textit{chunk-size} (C/P) play a critical role in determining how tokens are compressed using SSMs. The chunk-size refers to the number of consecutive tokens that are grouped together and processed as a single unit by the SSM to create a compressed key-value representation. The \textit{leading tokens} (L) specifies the number of recent tokens that should retain \textit{full attention} after the SSM. This is akin to sliding-window attention and has a constant cost as shown in Figure \ref{fig:leading_chunking}. Chunked attention with $L > 1$ would need us to parse the SSM block outputs as described in Equation \ref{eqn:leadingeg}.



\begin{equation} 
\begin{bmatrix} \mathbf{K}_{\text{SSM}} \\ \mathbf{V}_{\text{SSM}} \end{bmatrix} = 
\begin{bmatrix} 
\mathbf{K}^{(1)}[-1] & \cdots & \mathbf{K}^{(\frac{n}{P})}[-L:-1] & \mathbf{K}^{(\frac{n}{P})}[-1] \\ 
\mathbf{V}^{(1)}[-1] & \cdots & \mathbf{V}^{(\frac{n}{P})}[-L:-1] & \mathbf{V}^{(\frac{n}{P})}[-1] 
\end{bmatrix} 
\label{eqn:leadingeg}
\end{equation}

\textbf{Other Design Considerations: } In developing Attamba, several design choices were empirically validated, detailed in Appendix \ref{sec:appdx}. First, we found that removing Key-Value projection weights did not significantly impact model quality (1\% perplexity difference), simplifying the architecture. Secondly, cyclic chunk boundaries across layers mitigate bias introduced by fixed chunk boundaries on the input sequence (5\% improvement). Third, increasing SSM state dimensions beyond $D_{s} > 32$ yielded diminishing returns on $P=8$ ($<$ 1\% perplexity difference), allowing us to minimize SSM parameter overhead. Fourth, preserving leading tokens as un-chunked ensured improved model quality by maintaining full attention on recent tokens, emulating sliding window behavior (8.5\% improvement with $L = P$). Further, incorporating residual connections in Key-Value SSMs improved model quality, even without K-V projections. Finally, Attamba was robust to even \textbf{randomized chunk boundaries} (at both train and test time!). These are explained in more detail in the Appendix \ref{sec:appdx}.

\begin{figure*}[ht!]
        \centering
        \includegraphics[width=\linewidth]{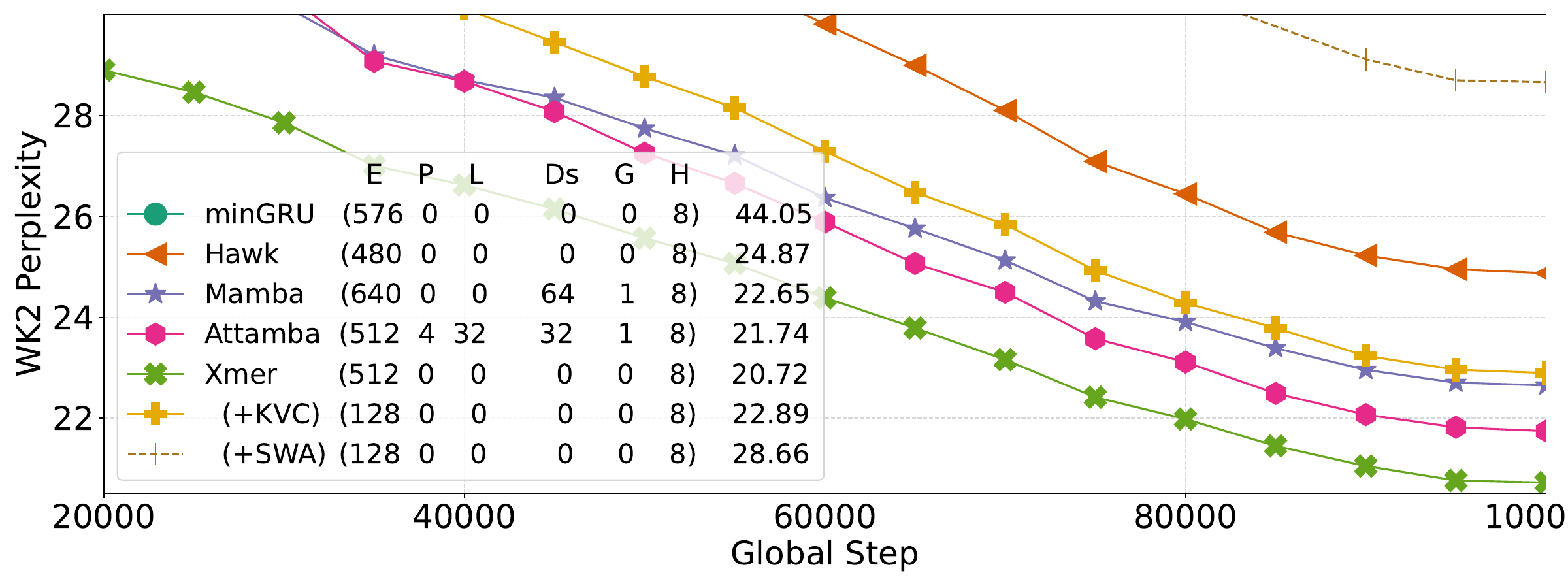}
        \caption{Comparing Attamba with SSMs (Mamba), minGRU, Hawk and Transformers (Xmer) by training on 8 billion tokens. \texttt{E, P, L, D$_s$, G, H} denote Model-Dim, Chunk Size, Leading Tokens, SSM State-Dim, Num. Groups and Num. Heads respectively, 0 when not applicable. Models $\in [60, 64]$M params, with Transformer having $4\times$ larger KV and attention map footprint. (+KV) \& (+SWA) transformer variants are 53M params, to match Attamba KV-Cache and Attention Map memory footprint more closely. \href{https://wandb.ai/akhauriyash/attamba_long}{[Logs]}}
        \label{fig:expt8}
\end{figure*}

\section{Experiments}

In this section, we present experimental results comparing the WikiText2 test-set perplexity. Training is done on 10\% of dclm-baseline-1.0 \cite{li2024datacomplm}, with a batch size of 16, sequence length of 1024. We use the Meta Lingua \cite{meta_lingua} framework. Unless otherwise specified, we train on approximately 1B tokens (982M tokens). \textit{Where relevant, we add the final WK2 perplexity in the graph legend}.

\subsection{On transformer baselines}
\label{subsec:tbaslines}

Attamba compresses the sequence length for keys and values, significantly reducing KV-Cache size and the operational intensity of the $L^2$ attention map, with the majority of savings occurring in inference-time activations. Comparing Attamba directly with a transformer of similar parameter count is not ideal, as traditional transformers incur much larger KV-Cache and attention map overhead. To provide a fairer context for Attamba's performance, we construct baselines with reduced KV-Cache sizes and attention map dimensions, detailed in Appendix~\ref{sec:appdx}. Specifically, for transformers, we emulate smaller KV-Cache sizes by reducing the attention model dimension $F$ such that $F = \frac{E}{P}$, and smaller attention maps by employing sliding window attention (SWA) during evaluation.

\begin{figure}[b!]
    \centering
    \includegraphics[width=\linewidth]{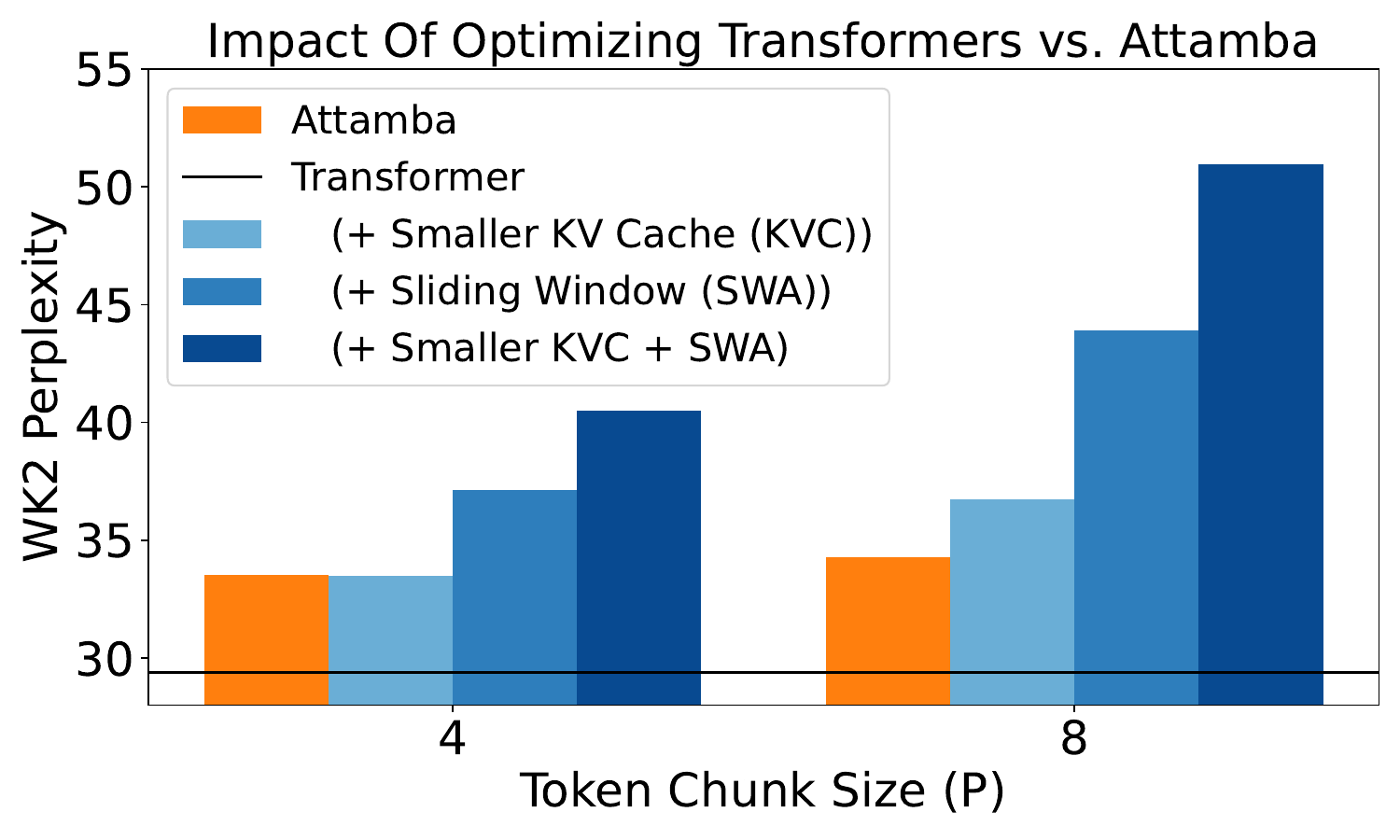}
    \caption{Comparing Attamba with a base transformer with matching parameter counts. Further, we train variants with smaller KV-Cache size to match Attamba. Additionally, to match the attention map size, we evaluate these models in \textit{Sliding Window} attention with window size $= \frac{L}{P}$. Attamba significantly out-performs a similarly sized transformer baseline (Smaller KVC + SWA). \href{https://wandb.ai/akhauriyash/attamba_isoStudy/}{[Logs]} 
    }
    \label{fig:iso_study}
\end{figure}

In Figure~\ref{fig:iso_study}, we observe that transformers with a $4\times$ to $8\times$ larger KV-Cache and attention map can outperform Attamba, but these comparisons do not reflect equivalent memory or computational constraints. When matched for KV-Cache and attention map size, Attamba consistently outperforms transformer baselines, showcasing its efficiency. Furthermore, as sequence length increases, transformers must reduce their attention dimension $F$ proportionally, resulting in greater trade-offs in quality. Attamba, on the other hand, demonstrates robust and scalable token compression, with a mere 2.2\% perplexity increase when transitioning from $P=4$ to $P=8$, highlighting its ability to balance efficiency and model quality effectively, particularly for long-context tasks.

\subsection{Comparison with Mamba, minGRU, Hawk} 

We conduct an extended training experiment with Attamba, Transformer (Xmer), minGRU \cite{mingru}, Hawk \cite{de2024griffin}, Mamba models within the parameter budget $\in [60,64]M$ params (roughly \textit{iso-Parameter} configurations), training for 100,000 steps over 8 billion tokens, as shown in Figure \ref{fig:expt8}. We also construct appropriate baselines for transformers, by reducing the model-dimension (specifically in the attention mechanism) to emulate KV-Cache compression (+KVC) and testing sliding window attention (+SWA), similar to Section \ref{subsec:tbaslines}. 

We find that Attamba out-performs fair transformer baselines as well as Mamba. Mamba will still show better performance scaling for extremely long context, but model quality may suffer ~\cite{wang2023stablessm_statecollapse}. Since Attamba uses SSMs to compress fixed-sized chunks of tokens, SSMs will not have to scale beyond their trained chunk-length, but merely attend to compressed token representations. Transformer (+KVC) is 5\% worse, but still materializes a $L^{2}$ attention map. Transformer (+SWA) is conducted on top of the Transformer (+KVC) variant, with a notable dip in perplexity due to a significantly constrained attention map.

\section{Limitations}

Currently, most of our training is limited to only 1B tokens on a 60M parameter model on a single A6000 GPU.  Further, our test-time evaluation is on WikiText2 \cite{merity2016pointer_wk2}, a task that is highly local. Thus, our variants Attamba with chunk-size 128 performs extremely well (Appendix \ref{sec:appdx}). While this variant offers a $128\times$ reduction in KV-Cache size and attention op-intensity over longer contexts, we also maintain full attention on the leading (latest) 128 tokens. This is why, it out-performs even Attamba with chunk size 4. From Figure \ref{fig:expt6}, we can see this in more detail, specifically, our Attamba P128 L1 variant (true $128\times$ KV-Cache reduction, with only 1 leading token (for self-attention)) performs significantly worse than Transformers. However, Attamba with chunk size 8 and 64 uncompressed leading tokens gives us a KV-Cost : $(\frac{L_{K} + L_{V}}{8} + 2\times64)E$ which offers $\approx 8\times$ KV-compression and $\approx 1/8$ attention computation cost with a 10\% perplexity trade-off. However, a transformer with model-dimension $E/8$ on attention performs $7.11\%$ worse, and has no memory savings on the attention map, as each head would still materialize the $L^{2}$ tensor. Further, our method leads to no improvements in the FFN, as we preserve the query sequence length to enable auto-regressive training of the transformer. Finding the right transformer design for a fair comparison is key to better understand trade-offs of attention on compressed states. Modifications in chunking strategy, or training on more tokens may alleviate issues with high chunk-sizes, but more thorough evaluation of this methodology on Long-Context evaluation, retrieval and other tasks are key to understand if effective attention can be achieved on compressed states. Finally, every model we have reported has been trained from scratch, exploring fine-tuning strategies and inference-time studies on chunk boundary robustness of models, as well as the impact of leading token $L$ (sliding window attention) need to be tested.

\newpage

\section{Conclusion}

Attamba introduces a novel approach to efficiently handle long sequences in transformers by integrating State-Space Models (SSMs) to compress tokens, reducing attention cost and KV-cache memory requirements. Experiments show that cyclic chunking outperforms other strategies, maintaining competitive performance with significant efficiency gains. By replacing conventional key-value projection matrices with SSMs and incorporating variable-length token chunking, Attamba effectively balances computational and memory efficiency, potentially enabling a smooth transition between quadratic and linear scaling if SSMs are flexible to chunk boundary lengths. However, our evaluation was limited to small-scale models and local tasks, such as WikiText2. Consequently, the observed performance improvements may not directly generalize to long-context benchmarks or billion-parameter language models. Future research should focus on extensive evaluation across a wider range of long-context tasks, exploration of token importance-based chunking strategies, and a deeper investigation into the trade-offs between efficiency and information retention in compressed states.

\bibliography{example_paper}
\bibliographystyle{icml2024}

\clearpage

\begin{figure}[ht!]
    \centering
    \includegraphics[width=0.8\linewidth]{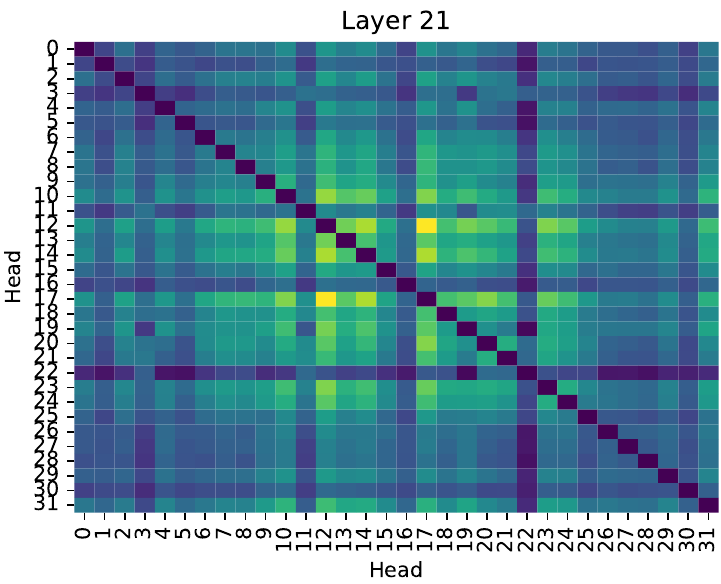}
    \caption{Each head in Llama-2-7B attends to tokens in a manner that is largely uncorrelated (Kendall-Tau $\in [-0.2, 0.8])$ with other heads.}
    \label{fig:tokvar}
\end{figure}

\begin{figure*}[ht!]
    \centering
    \includegraphics[width=\linewidth]{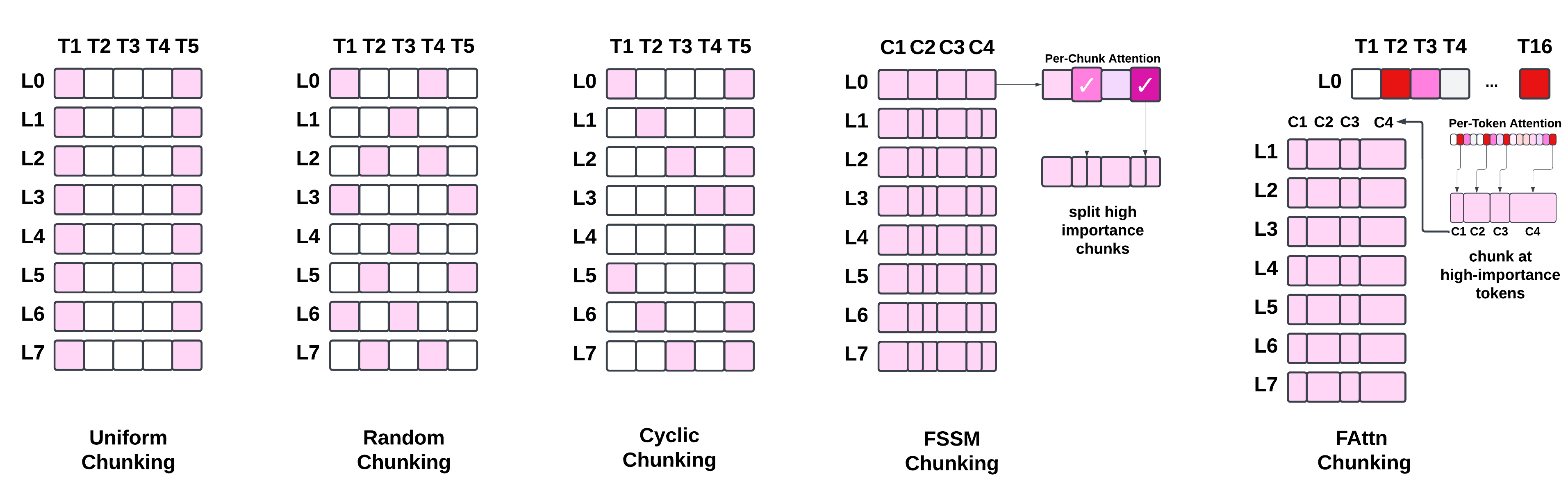}
    \caption{Different token-chunking strategies we investigate. \texttt{L, T, C} represent layer, token and chunk respectively. }
    \label{fig:chunking_strategies}
\end{figure*}
\appendix

\section{Appendix}
\label{sec:appdx}

\begin{figure*}[ht!]
    \centering
    \includegraphics[width=\linewidth]{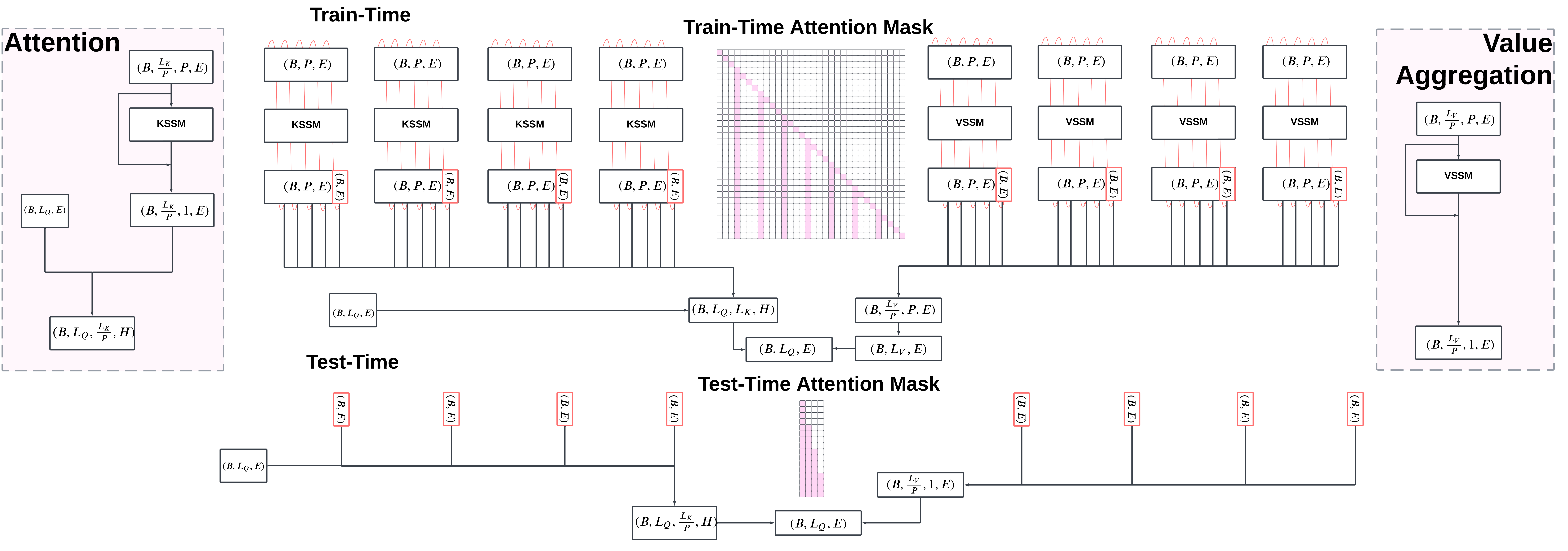}
    \caption{Attamba employs Key and Value State-Space Models (SSMs) to accumulate local information within chunks of tokens. At test time, only the final accumulated activations from each chunk are used in the standard attention mechanism. The red lines denote the auto-regressive SSMs, accumulating \textit{causally valid} local context within chunks. This approach significantly reduces attention complexity by compressing multiple tokens into single representations, while preserving essential contextual information from each chunk.}
    \label{fig:attamba_algo_appx}
\end{figure*}

\subsection{On Token Chunking}
\label{subsec:var_tok_chunk}

\begin{figure*}[t!]
    \centering
    \begin{minipage}[t]{0.48\textwidth}
        \centering
        \includegraphics[width=\linewidth]{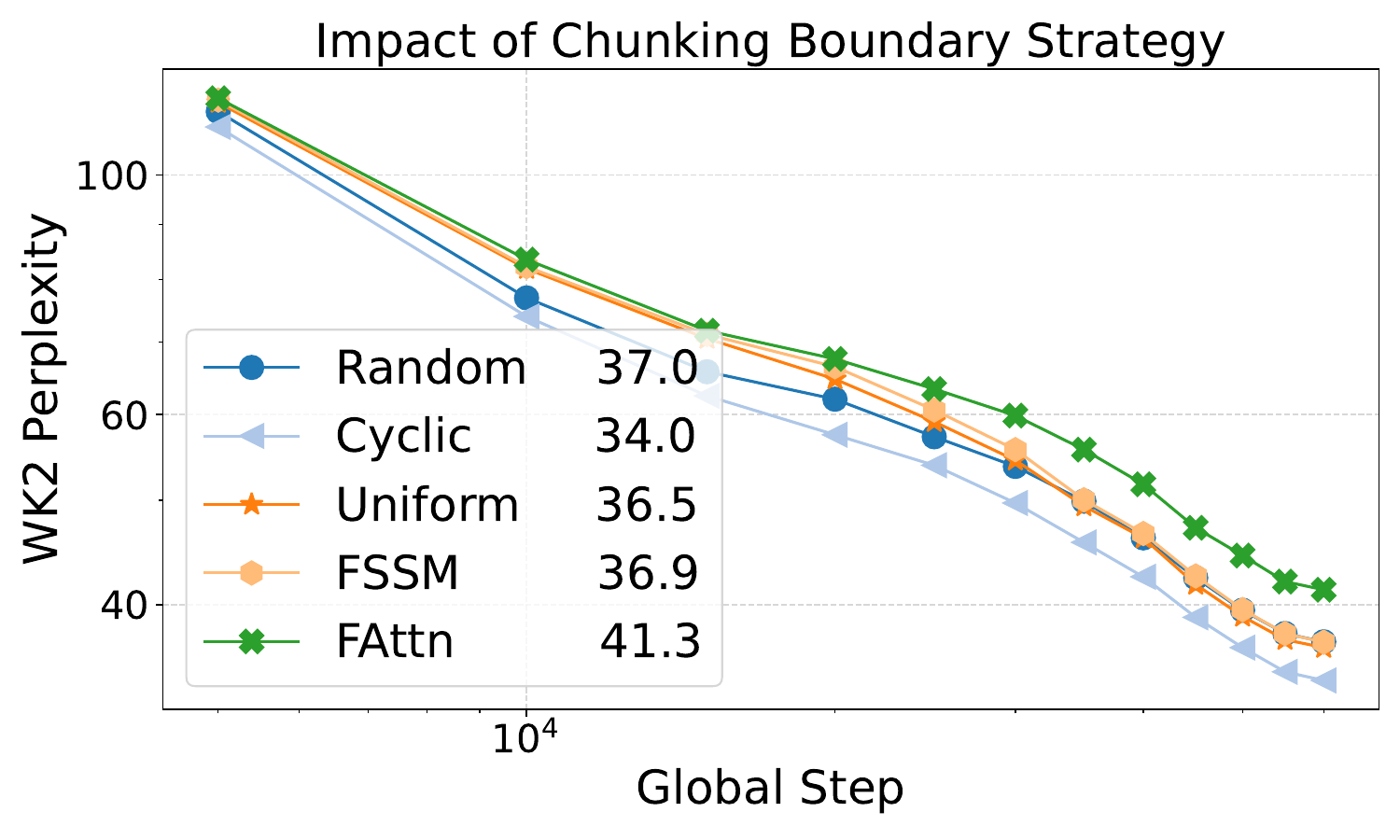}
        \caption{A simple cyclic chunk boundary performs better than other strategies. Notably, randomized chunk boundaries work as well as uniform chunking, indicating potential for flexibility in test-time token chunking.}
        \label{fig:expt3}
    \end{minipage}
    \hfill
    \begin{minipage}[t]{0.48\textwidth}
        \centering
        \includegraphics[width=\linewidth]{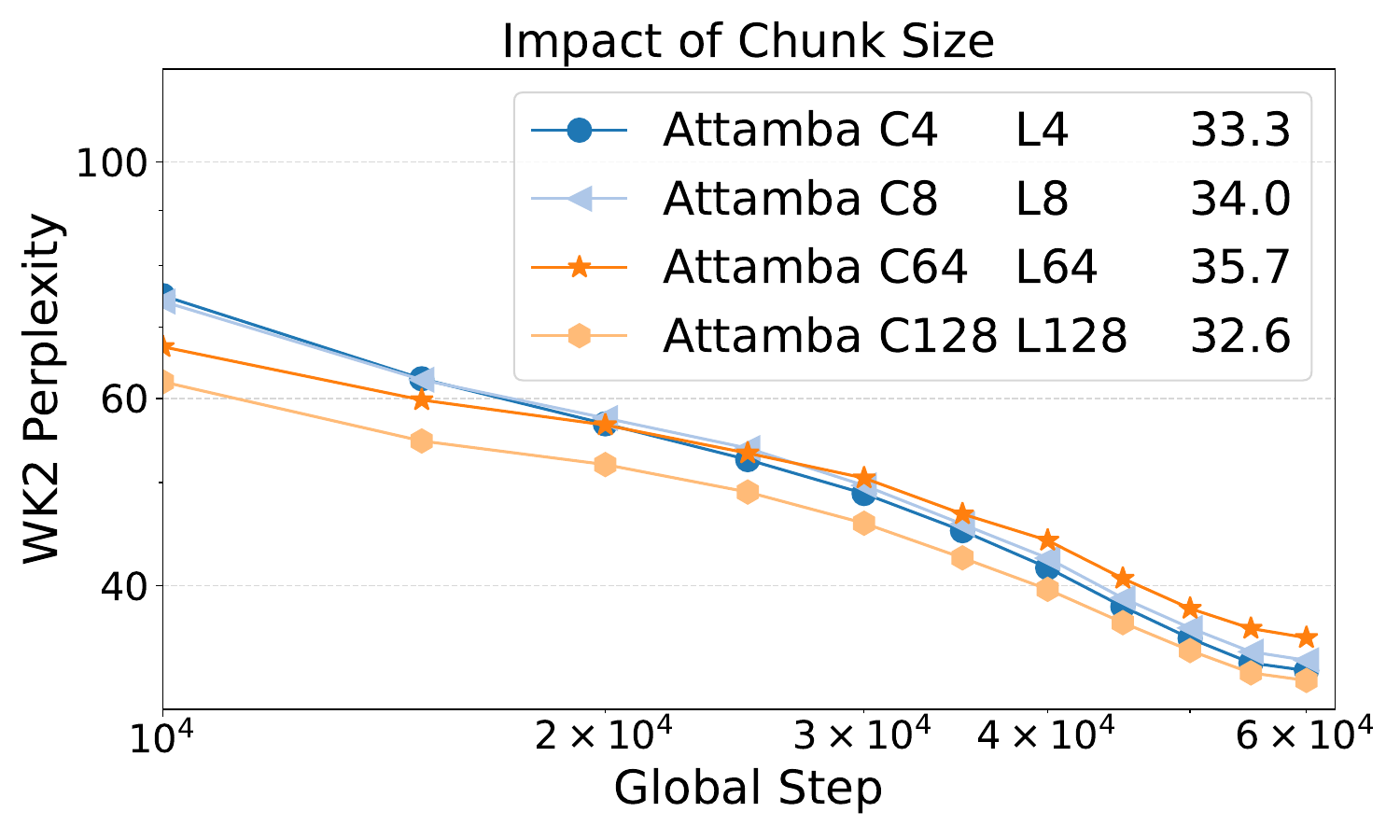}
        \caption{Chunk size of 128 implies a $~128\times$ smaller KV-Cache. It outperforms Chunk 4/8/64 because we do full-attention on partial-chunks, giving significant advantage as chunk-size increases on local evaluation tasks like WikiText2.}
        \label{fig:expt4}
    \end{minipage}
    \label{fig:side_by_side_expts_2}
\end{figure*}

Processing sequences in fixed-size chunks simplifies implementation, but can limit models flexibility. Prior research \cite{zhang2023h2o} has found that certain tokens contribute largely to the perplexity and are contextually important~\cite{shadowllm}. In this context, having chunk boundaries at \textit{important} tokens for a given query can improve model quality, and maintaining this flexibility for research in token importance prediction can unlock improved efficient language modeling. To enable efficient processing of sequences with arbitrary chunk boundaries in the SSM, we do not reshape or explicitly chunk the sequence. Instead, we utilize the \texttt{cu\_seqlens} tensor in the Mamba library. This allows us to handle variable-length chunk boundaries without padding overhead. Figure \ref{fig:chunking_strategies} depicts token-chunking strategies we try. Random-Chunking partitions the sequence into $P$ chunks with sizes $\{s_i\}_{i=1}^{P}$, where $s_i \sim \text{Random}(S)$ and $\sum_{i=1}^{P} s_i = n$. From Figure \ref{fig:expt3}, we can see that Random-Chunking works as well as Uniform-Chunking, indicating that SSM based token chunking is flexible.

\begin{figure*}[h]
    \centering
    \begin{minipage}[t]{0.48\textwidth}
        \centering
        \includegraphics[width=\linewidth]{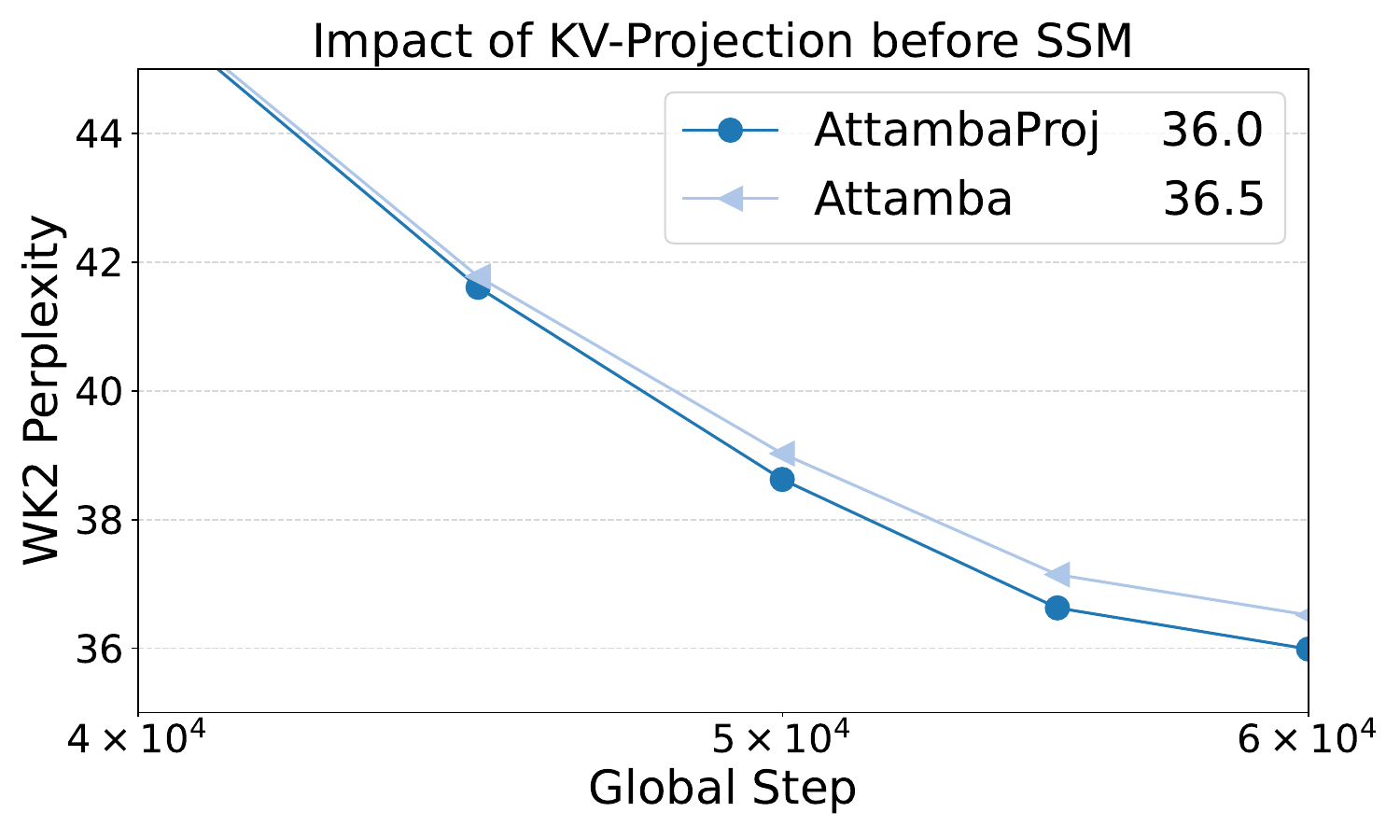}
        \caption{Removing the Key-Value projection matrices when using K-V SSMs does not impact WikiText2 test-perplexity significantly.}
        \label{fig:expt1}
    \end{minipage}
    \hfill
    \begin{minipage}[t]{0.48\textwidth}
        \centering
        \includegraphics[width=\linewidth]{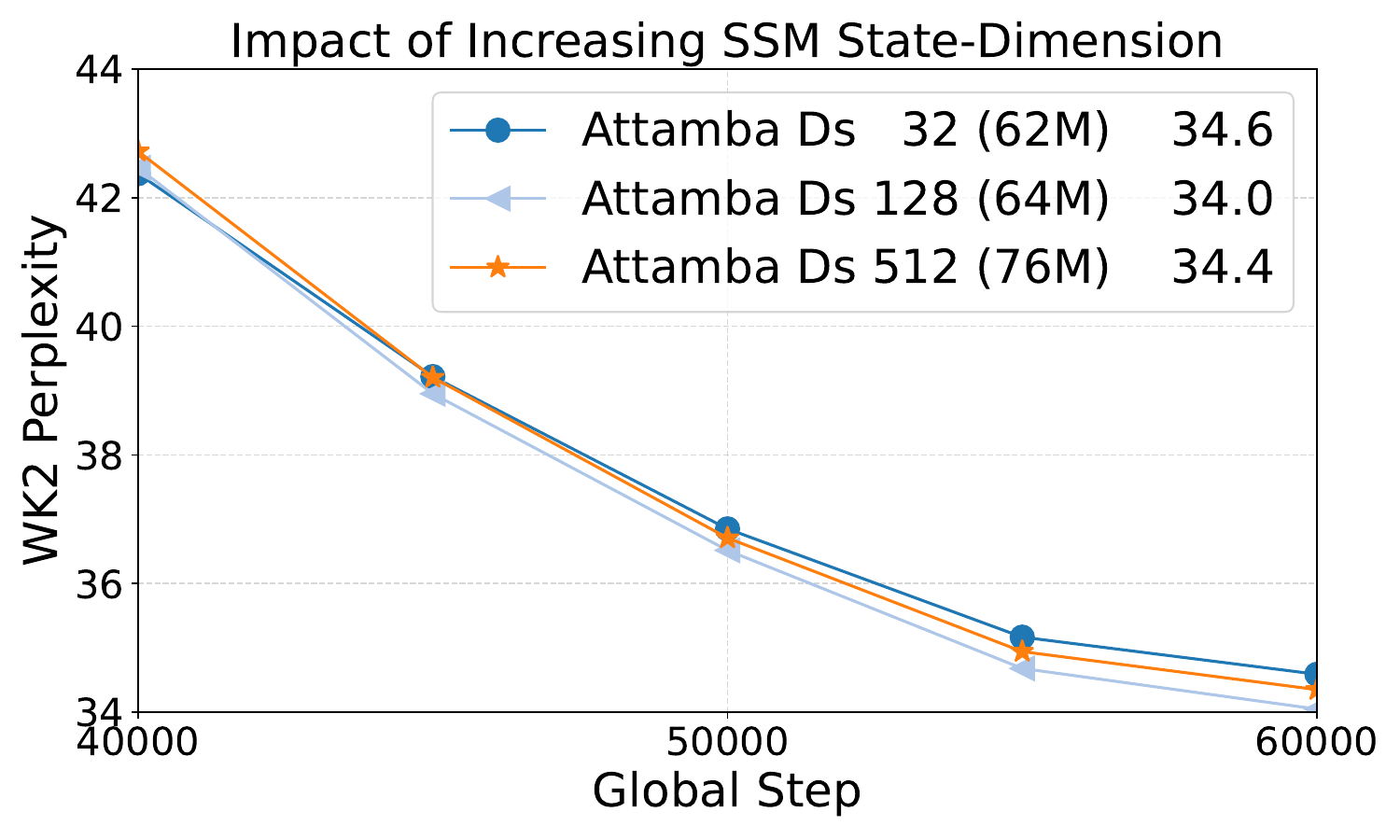}
        \caption{Increasing the state dimension ($D_s$) of Key-Value SSMs does not improve perplexity when processing chunks of 8 tokens.}
        \label{fig:expt2}
    \end{minipage}
    \label{fig:side_by_side_expts}
\end{figure*}

\begin{figure*}[ht!]
    \centering
    \includegraphics[width=\linewidth]{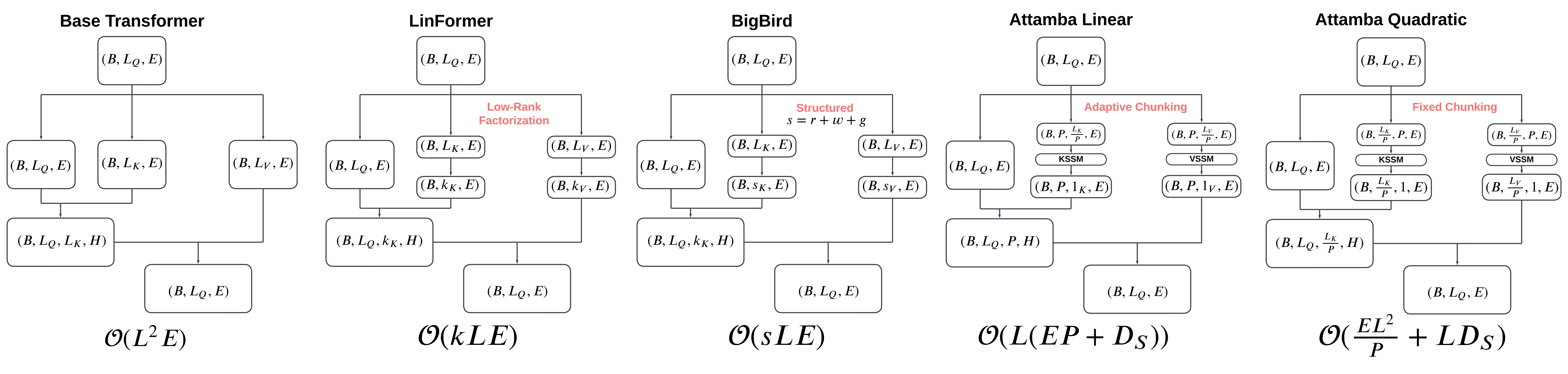}
    \caption{Attamba-Linear maintains linear complexity, by having a fixed-size attention, and dividing the sequence length ($L$) into chunks. Attamba-Quadratic has quadratic complexity (albeit lower FLOPs/Memory than standard transformer) as the SSM only processes $P$ tokens. \texttt{w, r, g, k, E} denote window, random, global, low-rank dimension and model dimension respectively.}
    \label{fig:attamba_variants}
\end{figure*}

\subsubsection{Cyclic Chunking}

Fixed chunk boundaries intorduce biases into the model, as tokens near chunk boundaries may be over-represented due to their position. To aim to mitigate this, we employ a cyclic chunking strategy, with different layers using chunk boundaries with a layer offset. Essentially, the chunk boundary is shifted by the index of the current layer. This ensures different layers process different token groupings, distributing boundary effects across the model. 

By varying chunk boundaries across layers we encourage the SSM to be robust to chunk boundaries. We experiment with more chunk boundary decision strategies detailed in the next subsection, but find that cyclic chunking is a simple and effective strategy. 

\begin{figure*}[ht!]
    \centering
    \begin{minipage}[t]{0.48\textwidth}
        \centering
        \includegraphics[width=\linewidth]{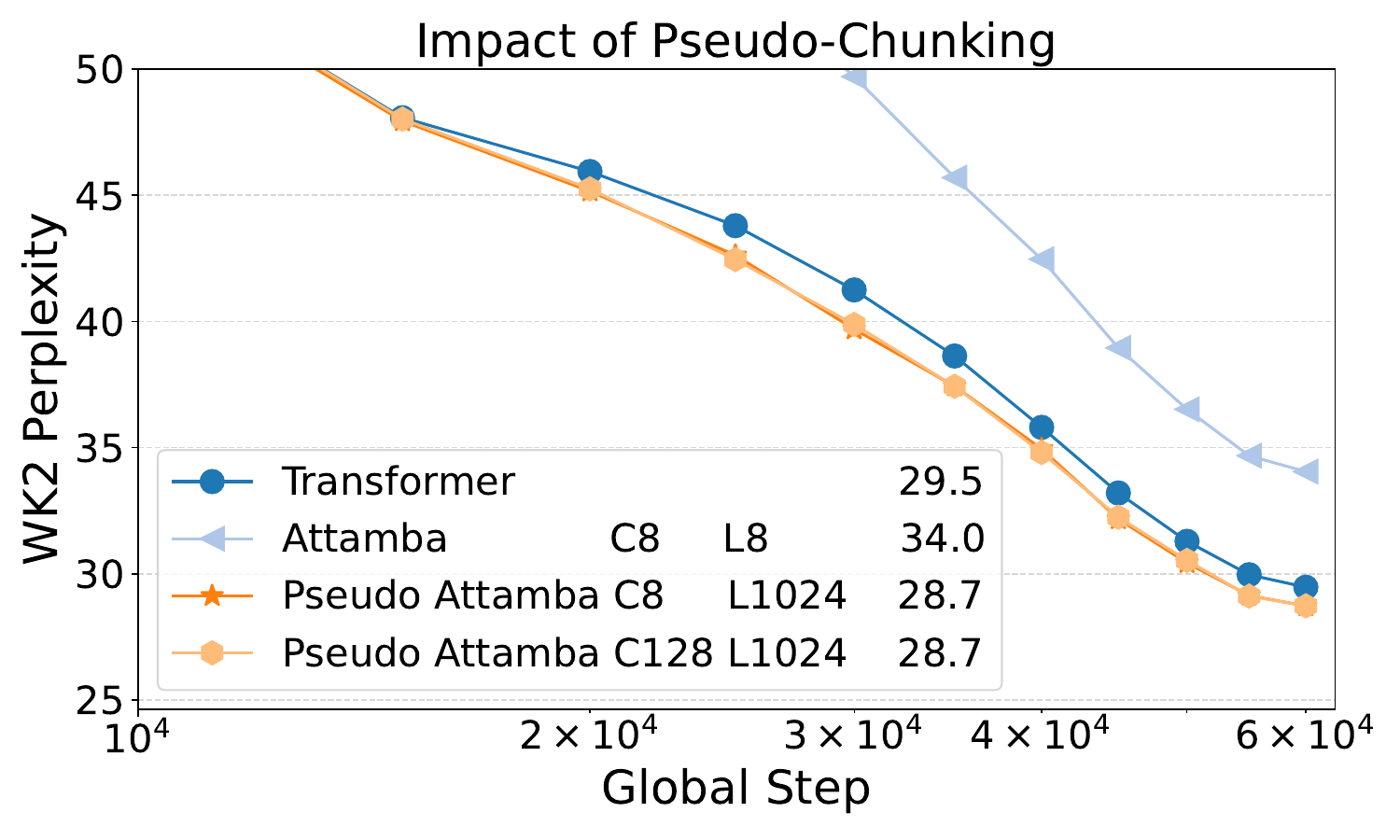}
        \caption{Pseudo-Chunking (replacing Key-Value projection matrices with SSMs, but attending to all tokens) can marginally improve transformer perplexity. (C: Chunk Size)}
        \label{fig:expt5}
    \end{minipage}
    \hfill
    \begin{minipage}[t]{0.48\textwidth}
        \centering
        \includegraphics[width=\linewidth]{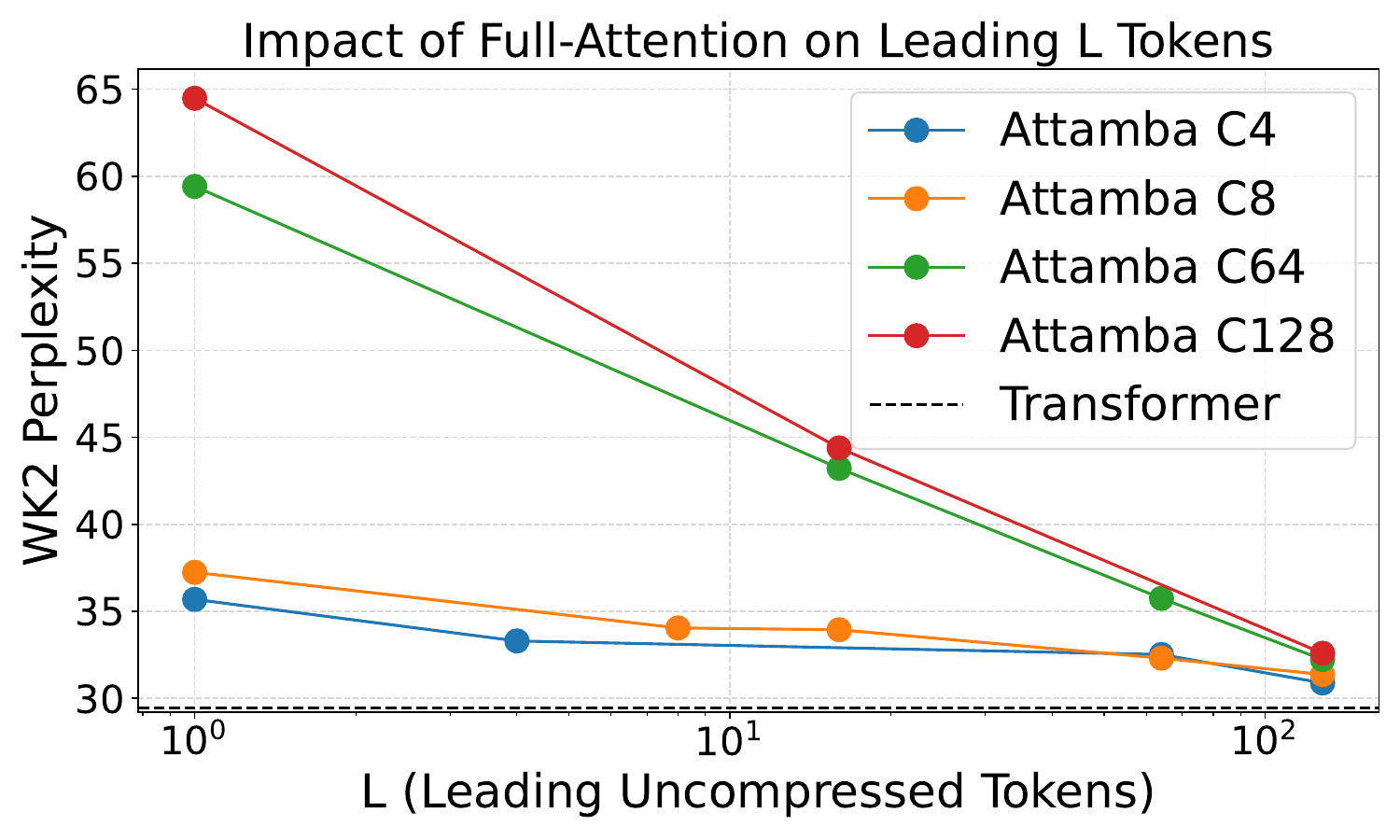}
        \caption{Leading tokens improve test-time perplexity, a proper chunk-size to leading token trade-off is important. This may also indicate limitations in Attamba's ability to compress tokens.}
        \label{fig:expt6}
    \end{minipage}
    \label{fig:side_by_side_expts_2}
\end{figure*}

\begin{figure*}[ht!]
    \centering
    \begin{minipage}[t]{0.48\textwidth}
        \centering
        \includegraphics[width=\linewidth]{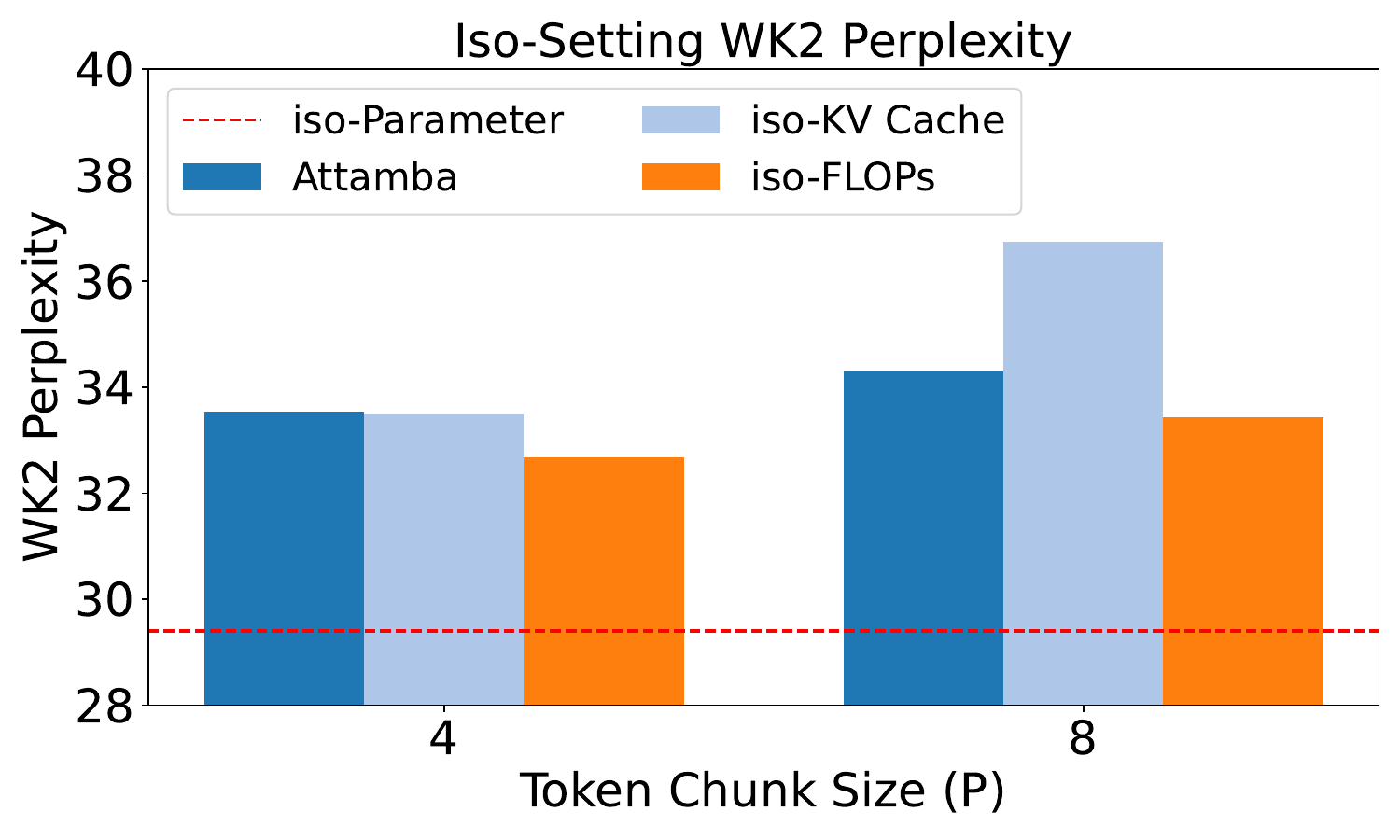}
        \caption{iso-Parameter and iso-FLOPs still has higher memory overhead and does not address the $L^{2}$ attention and KV-Cache overhead.}
        \label{fig:iso_all}
    \end{minipage}
    \hfill
    \begin{minipage}[t]{0.48\textwidth}
        \centering
        \includegraphics[width=\linewidth]{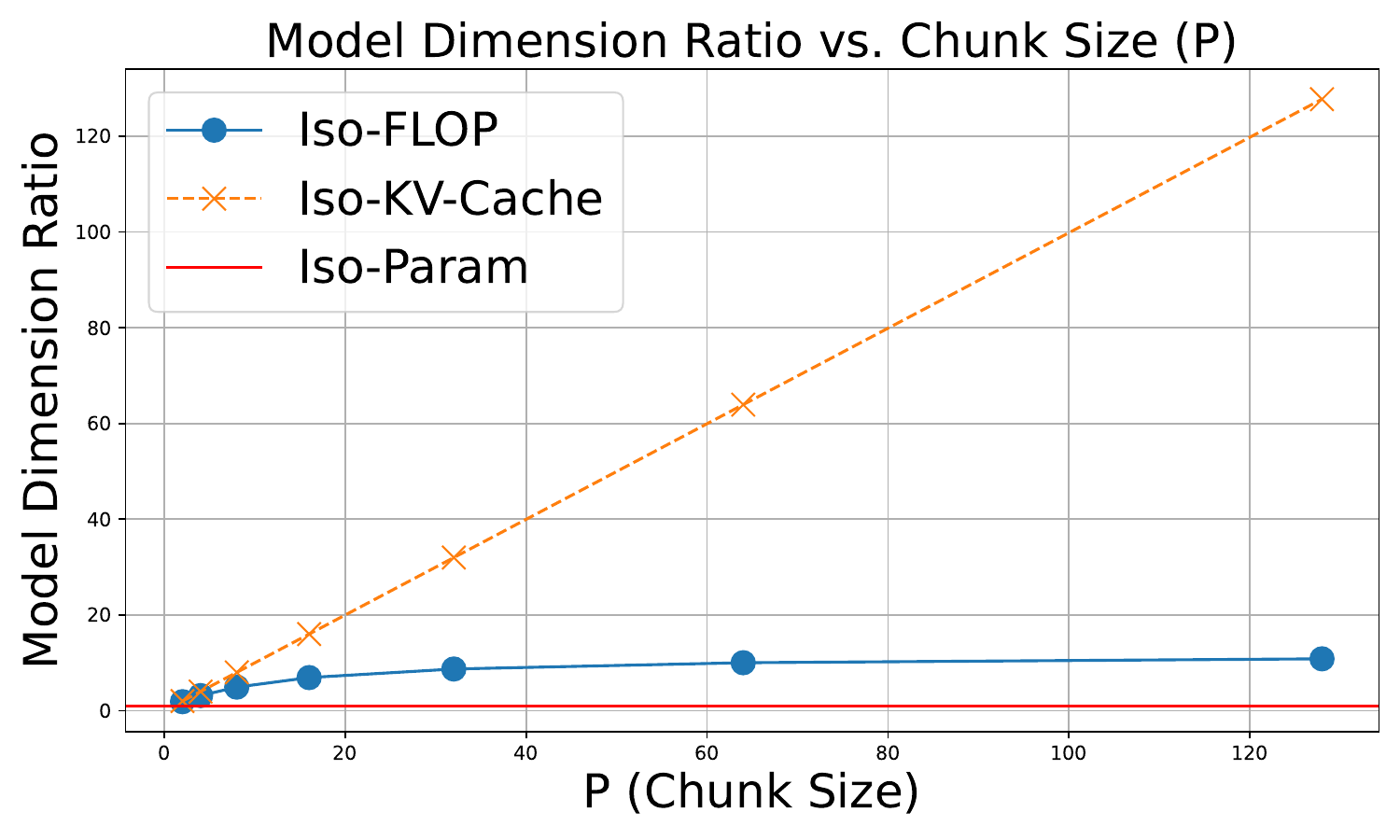}
        \caption{The ratio of Attamba Model Dimension with Transformer Attention Model Dimension ($E$) required for varying iso-setting baselines as we scale chunk size.}
        \label{fig:P_vs_msize}
    \end{minipage}
    \label{fig:side_by_side_expts_2}
\end{figure*}

\subsubsection{On chunk boundary selection}

In addition to cyclic chunking, we explore alternative strategies for determining chunk boundaries to improve model performance. One such method, referred to as \textbf{FAttn}, involves using full attention in the first layer to identify important tokens based on attention magnitudes. Specifically, we compute the attention weights in the first layer using the standard full attention mechanism and select the sequence positions with the highest attention scores as chunk boundaries for subsequent layers. This aims to place chunk boundaries at tokens deemed important by the model, potentially enhancing the quality of the compressed representations.

Another approach, termed \textbf{FSSM}, utilizes the attention map of chunks from the first layer with uniform chunk boundaries. We compute the attention scores for each chunk and identify the top $k$ chunks with the highest attention values. These selected chunks are then split into 2 smaller chunks in the subsequent layers, effectively allocating more resources to the most informative parts of the sequence.

While we experiment with an array of chunk boundary selection methods, we found that cyclic chunk boundaries yield the best quality improvements. On the other hand, \textbf{FSSM} and \textbf{FAttn} do not aid chunk boundary selection too much. This may be attributed to our finding that different heads attend to different tokens, and using the first layer to decide all head-boundaries is worse than randomized/cyclic methods. This effect is visible even within a single layer on the Llama-2-7B model, in Figure \ref{fig:tokvar} we can see that for 1024 token context on WikiText2, each head on layer 21 has low correlation between tokens attended to. 

\subsection{On Pseudo-Chunking}

We find that SSMs can serve as a drop-in replacement for the key-value projection matrices, enabling us to save on KV-Cache and the quadratic attention cost by token chunking. However, we can also \textit{pseudo-chunk} the input. That is, given a parameter budget for model size, we can use the SSM as a replacement for projection matrix, and maintain full-attention. This is more computationally expensive, but also improves model quality. Psuedo-chunking can be thought of as Attamba, where \textbf{L} (Leading Tokens in Figure \ref{fig:leading_chunking}) is the same as the sequence length.

\section{Experiments}

In this section, we present experimental results comparing the WikiText2 test-set perplexities during model training for a 60M parameter transformer model, with 8 layers, 8 heads and 512 model-dimension on a single A6000 GPU. Training is done on 10\% of dclm-baseline-1.0 \cite{li2024datacomplm}, with a batch size of 16, sequence length of 1024. We use the Meta Lingua \cite{meta_lingua} framework. Unless otherwise specified, we train on approximately 1B tokens (982,630,400 tokens). \textit{Where relevant, we add the final WK2 perplexity in the graph legend}.

\textbf{SSMs For Key-Value Projections: }
We replace the KV projection matrices with SSMs to enable chunked-attention. In Figure \ref{fig:expt1}, we compare the WikiText2 perplexities. We use uniform 8-token chunking and compare models with and without KV-weight-projections. We find marginal benefits in perplexity by keeping the KV projection matrices before the SSM, and decide to remove it. This also reduces the parameter count and overall FLOPs of the model.

\textbf{SSM Parameter Count: }
The SSMs need to do the Key-Value projections, but also compress states for accurate attention, as well as information propagation in the value activations. Thus, the hidden-state of the SSM is important. In Figure \ref{fig:expt2}, we study the impact of varying SSM size, from total approximate parameter-overhead of 2M, 4M and 16M parameters on a 60M parameter model. We see that for a token-chunking size of 8, the SSM does not need to be too large, as the benefit is marginal. For the rest of the experiments, we keep the total SSM parameter overhead 4M, but this can likely be optimized with chunk-size.

\textbf{Chunking Methodology: }
Chunking can significantly impact model quality. To test it, we try different chunking methodologies \textit{Uniform}, \textit{Random}, \textit{Cyclic}, \textit{FAttn} and \textit{FSSM}. From Figure \ref{fig:expt3}, we can see that cyclic performs the best. However, it is important to note that \textit{Random} chunking performs similarly to \textit{Uniform} chunking, indicating that Attamba is robust to chunking boundaries, and can significantly benefit from research in token importance prediction. 

\textbf{Token Chunking Size: }
As shown in Figure \ref{fig:leading_chunking}, our chunking methods keeps full attention on the final chunk by default (leading tokens smaller than the chunk size are preserved). This means that as we increase token chunking size, latest \texttt{chunk\_size} tokens get full attention. This is not compulsory, but we aim to emulate the local sliding window attention with this, as the computational over-head is constant. In Figure \ref{fig:expt4}, we compare different chunk sizes. We observe a trend where smaller chunk sizes yield better performance, with Chunk 4 outperforming Chunk 8, which outperforms Chunk 64. However, Chunk 128 performs the best, this is simply because WikiText2 is a highly local task, and keeping the latest 128 tokens un-chunked improves perplexity. More rigorous long-context evaluation is required to determine how well token-information is preserved.

\textbf{Pseudo-Chunking: }
We replace the KV projection weights with SSMs, and enforce chunk boundaries in the attention mask to emulate KV-Cache optimizations. However, it is also possible to use the SSM so that each token has more information about prior local tokens, without optimizing the transformer for performance. This can be achieved by simply keeping a purely causal mask on Attamba, with no chunk-boundaries. In Figure \ref{fig:expt5}, we find that pseudo-chunking can actually improve transformer performance, even in iso-parameter count settings.

\begin{figure*}[t!]
        \centering
        \includegraphics[width=\linewidth]{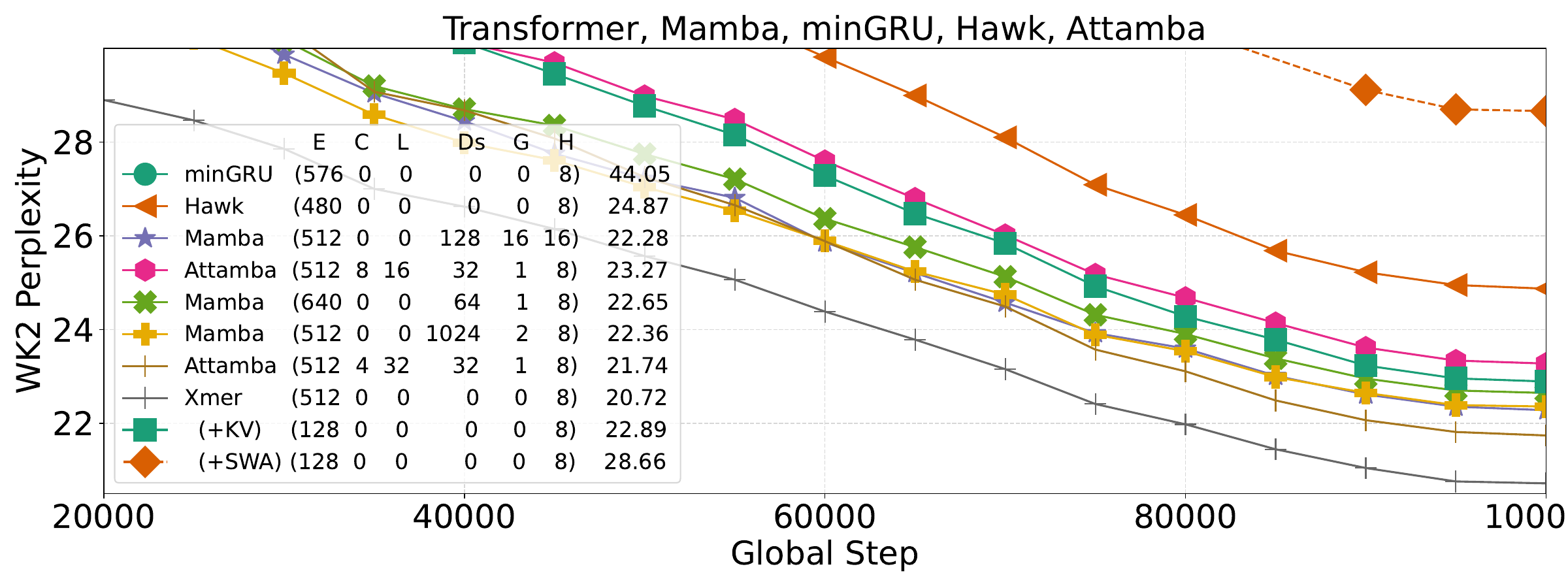}
        \caption{Comparing Attamba with SSMs (Mamba), minGRU, Hawk and Transformers (Xmer) by training on 8 billion tokens. \texttt{E, C, L, D$_s$, G, H} denote Model-Dim, Chunk-Size, Leading-Tokens, SSM State-Dim, Num. Groups and Num. Heads respectively, 0 when not applicable. Models $\in [60, 64]$M params, with Transformer having significantly larger KV-footprint \href{https://wandb.ai/akhauriyash/attamba_long}{[Logs]}} 
        \label{fig:expt8_longfull}
\end{figure*}

\textbf{Estimating FLOPs, KVCache and Activation Overhead}

Attamba compresses states differently from existing methods of controlling transformer architectures via model dimensions. Comparing Attamba solely with iso-parameter count baselines is inappropriate because transformers produce significantly larger intermediate activations, such as attention maps. To find appropriate transformer baselines, we use a simplified approach to calculate iso-KV-cache size, iso-memory, and iso-FLOPs settings for the \textit{Transformer Block}. These calculations exclude scaling, normalization, and softmax considerations, focusing on high sequence lengths.

We define the following parameters: Transformer attention-only model dimension (\(F\)), Attamba model dimension (\(E\)), number of heads (\(H\), assumed to be 1 unless otherwise stated), chunk size (\(P\)), sequence length (\(L\)), SSM dimension (\(D_S\)), and batch size (\(B\)). To find the right $F$, we solve simply by substituting the default Attamba configurations, and use this F dimension in the attention mechanism of the base-transformer.

\textbf{Iso-KV Settings: } For iso-KV settings, the appropriate $F$ is solved for as follows:
\begin{equation}
    2BLF = \frac{2BLE}{P} + 2BD_S
\end{equation}

\textbf{Iso-FLOPs Settings: } For iso-FLOPs settings, the appropriate $F$ is solved for as follows:
\begin{equation}
\begin{aligned}
    6BLF^2 + 4BL^2F = & \, 2BLE^2 \\
    & + 2BL\left(\frac{E}{H}(5HD_S + D_S) + 21D_S\right) \\
    & + \frac{4BL^2E}{P}
\end{aligned}
\end{equation}

This derivation is more verbose than Figure \ref{fig:attamba_complexity}, which included simplified equations for brevity. These formulations enable comparisons across iso-KV-cache, iso-memory, and iso-FLOPs scenarios.

\textbf{Iso-Activation Settings: } For iso-activation settings, the appropriate $F$ is solved for as follows:
\begin{equation}
    4BLF = 2BLE \left(1 + \frac{1}{P} \right) + 2BD_{S} + BL^{2}H\left(\frac{1}{P} - 1\right)
\end{equation}

Due to the $\frac{1-P}{P}$ term always being negative, and the quadratic $L^{2}$ scaling on high sequence lengths, we are unable to find an appropriate iso-activation transformer design in our budget. This is largely because Attamba significantly optimizes the $L^{2}$ attention mechanism, which reduces the activation footprint.

\subsection{Baselines}  

\begin{table}[t]
    \centering
    \begin{tabular}{cccccc}
    \toprule
    P & Attamba & IsoParam & IsoFLOP & IsoKV \\
    \midrule
    4 & 512 & 512 & 160 & 128 \\
    8 & 512 & 512 & 104 & 64 \\
    \bottomrule
    \end{tabular}
    \caption{Setting for Transformer Baseline (Model Dimension) for IsoFLOP and IsoKV at Fixed Attamba Dimension (E=512). Calculated for Sequence Length 4096.}
    \label{tab:iso_fkv}
\end{table}

\textbf{Iso-KV Baseline:} For iso-KV settings, the transformer model dimension is adjusted to equate the total KV-cache footprint with that of Attamba. This comparison highlights the memory savings achieved by Attamba's reduced KV-cache size, however \textbf{this baseline does not account for the $L^{2}$ attention matrix that is materialized.} In this sense, Attamba will still be significantly more efficient for long-context. For instance, at \(P = 4\), Attamba achieves the same KV-Cache size, but materializes a $4\times$ smaller attention map per-head.

\textbf{Iso-FLOPs Baseline:} Iso-FLOPs baselines align the computational cost of the transformer with Attamba by scaling down the transformer model dimension (\(F\)) to match FLOP counts as estimated by us in Appendix \ref{sec:appdx}. As demonstrated in Figure \ref{fig:P_vs_msize} and Table \ref{tab:iso_fkv}, this compares the efficiency of Attamba in scenarios where computational budgets are fixed. However, this also fails to account for the KV-Cache overhead and larger attention map.



\textbf{Iso-Parameter Baseline:} Here, transformer baselines are chosen such that their parameter count approximately matches Attamba. This comparison does not factor in differences in KV-cache size and attention computation but offers a straightforward view of the representational capacity of the models. 

Inference efficiency strongly favors Attamba due to reduced memory bandwidth requirements, a major bottleneck in transformers. Iso-KV baselines ignore the quadratic scaling of attention maps, Iso-FLOPs and Iso-Parameter baselines do not optimize for KV-cache or activation footprint.

As shown in Figure \ref{fig:iso_study}, Attamba consistently outperforms Iso-FLOPs models due to its ability to compress and operate on compressed tokens effectively. It performs similarly to Iso-KV models but achieves additional gains by reducing attention map operations, which scale quadratically with sequence length. This gap widens at higher sequence lengths (e.g., L $\geq$ 4096), where Iso-KV models require progressively smaller attention dimensions to match Attamba's efficiency. As is expected, we perform worse than iso-Parameter models but are significantly better on FLOPs, KV-cache size, and attention map efficiency.

\end{document}